\documentclass[10pt,twocolumn,letterpaper]{article}

\usepackage[pagenumbers]{cvpr} 
\usepackage[accsupp]{axessibility} 

\usepackage{times}
\usepackage{epsfig}
\usepackage{graphicx}
\usepackage{caption}
\usepackage{amsmath}
\usepackage{amssymb}
\usepackage{mathtools}
\usepackage{microtype}
\usepackage{multirow}

\usepackage{algorithm}
\usepackage{algorithmic}

\usepackage[dvipsnames]{xcolor}

\definecolor{cvprblue}{rgb}{0.21,0.49,0.74}
\usepackage[pagebackref,breaklinks,colorlinks,citecolor=cvprblue]{hyperref}
\usepackage{url}

\def\paperID{1002} 
\def\confName{CVPR}
\def\confYear{2024}

\title{An Edit Friendly DDPM Noise Space:\\Inversion and Manipulations}

\author{Inbar Huberman-Spiegelglas \qquad Vladimir Kulikov \qquad Tomer Michaeli\\
Technion  -- Israel Institute of Technology\\
{\tt\small inbarhub@gmail.com, \{vladimir.k@campus, tomer.m@ee\}.technion.ac.il}
}

\begin{document}
\twocolumn[{%
	\maketitle
	\vspace{-0.75cm}
	\renewcommand\twocolumn[1][]{#1}%
	\begin{center}
		\centering	
\captionsetup{type=figure}
\includegraphics[width=\textwidth]{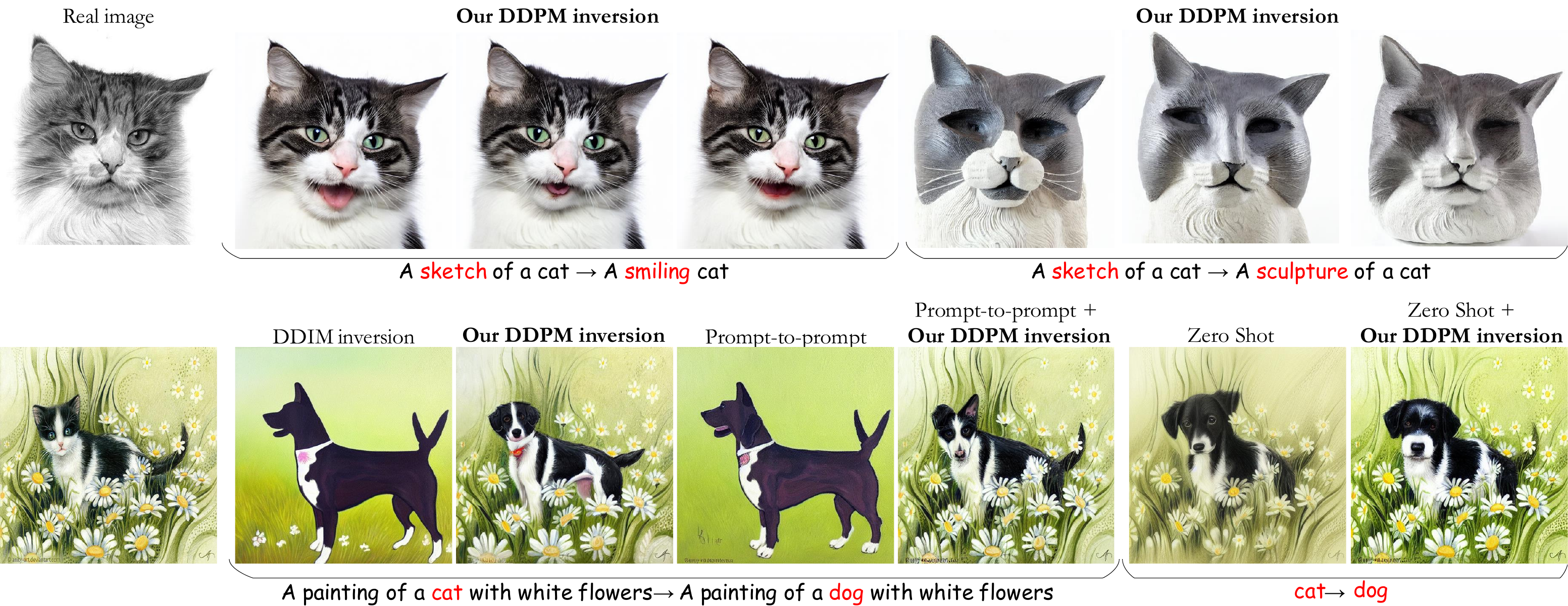}
\caption{\textbf{Edit friendly DDPM inversion.} We present a method for extracting a sequence of DDPM noise maps that perfectly reconstruct a given image. These noise maps are distributed differently from those used in regular sampling, and are more edit-friendly. Our method allows diverse editing of real images without fine-tuning the model or modifying its attention maps, and it can also be easily integrated into other algorithms (illustrated here with Prompt-to-Prompt \cite{Hertz22} and Zero-Shot I2I \cite{Parmar23})
}
\label{fig:teaser}
\end{center}%
}]
\begin{abstract}
Denoising diffusion probabilistic models (DDPMs) employ a sequence of white Gaussian noise samples to generate an image. In analogy with GANs, those noise maps could be considered as the latent code associated with the generated image. However, this native noise space does not possess a convenient structure, and is thus challenging to work with in editing tasks.
Here, we propose an alternative latent noise space for DDPM that enables a wide range of editing operations via simple means, and present an inversion method for extracting these edit-friendly noise maps for any given image (real or synthetically generated). As opposed to the native DDPM noise space, the edit-friendly noise maps do not have a standard normal distribution and are not statistically independent across timesteps. However, they allow perfect reconstruction of any desired image, and simple transformations on them translate into meaningful manipulations of the output image  (\eg shifting, color edits). Moreover, in text-conditional models, fixing those noise maps while changing the text prompt, modifies semantics while retaining structure. We illustrate how this property enables text-based editing of real images via the diverse DDPM sampling scheme (in contrast to the popular non-diverse DDIM inversion). We also show how it can be used within existing diffusion-based editing methods to improve their quality and diversity. 
Code and examples are available on the project's 
\href{https://inbarhub.github.io/DDPM\_inversion}{webpage}.


\end{abstract}

\section{Introduction}
\label{intro}

Diffusion models have emerged as a powerful generative framework, achieving state-of-the-art quality on image synthesis~\cite{Ho20,balaji22,Rombach22,pmlr22,saharia22,Ramesh22}. Recent works harness diffusion models for various image editing and manipulation tasks, including text-guided editing~\cite{Hertz22,Guillaume22,Narek22,avrahami22,Kim22}, inpainting~\cite{Lugmayr22}, and image-to-image translation~\cite{Saharia21,Meng22,Wu22}. A key challenge in these methods is to leverage them for editing of \emph{real} content (as opposed to model-generated images). 
This requires inverting the generation process, namely extracting a sequence of noise vectors that would reconstruct the given image if used to drive the reverse diffusion process.

Despite significant advancements in diffusion-based editing, inversion is still considered a major challenge, particularly in the denoising diffusion probabilistic model (DDPM) sampling scheme~\cite{Ho20}. Many recent methods (\eg \cite{Hertz22,Mokady22,Narek22,Guillaume22,Bram22,Parmar23}) rely on an approximate inversion method for the denoising diffusion implicit model (DDIM) scheme~\cite{Song21}, which is a deterministic sampling process that maps a single initial noise vector into a generated image. However this DDIM inversion method becomes accurate only when using a large number of diffusion timesteps (\eg 1000), and even in this regime it often leads to sub-optimal results in text-guided editing~\cite{Hertz22,Mokady22}. To battle this effect, some methods fine-tune the diffusion model based on the given image and text prompt~\cite{Bahjat22,Kim22,Valevski22,zhang23}. Other methods intervene in the generative process in various ways, \eg by injecting the attention maps derived from the DDIM inversion process into the text-guided generative process~\cite{Hertz22,Parmar23,Narek22,cao23}.

\begin{figure}
\centering
\includegraphics[width=\columnwidth]{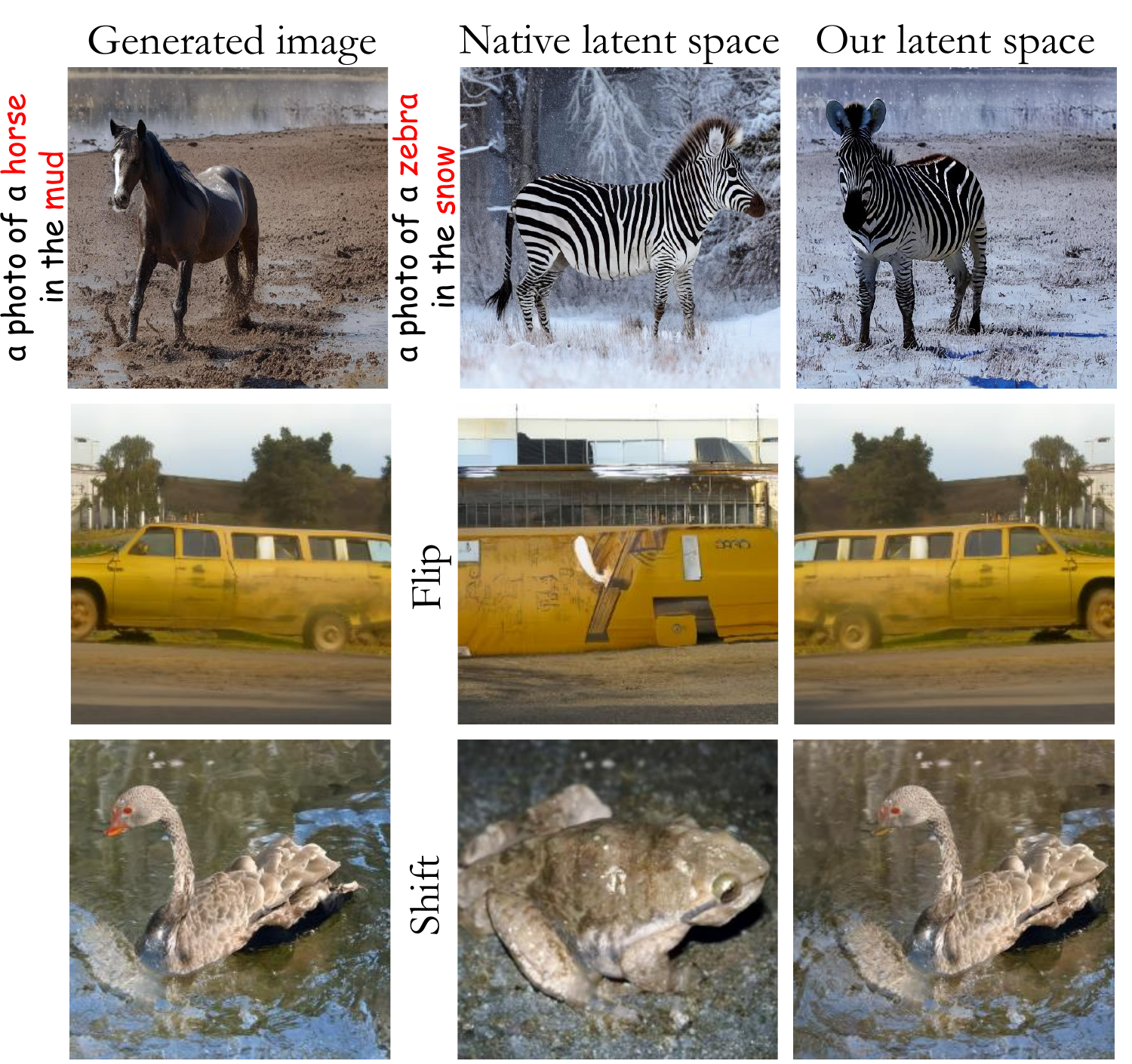}
\caption{\textbf{The native and edit friendly noise spaces.} When sampling an image using DDPM (left), there is access to the ``ground truth'' noise maps that generated it. This native noise space, however, is not edit friendly (2nd column). For example, fixing those noise maps and changing the text prompt, changes the image structure (top). Similarly, flipping (middle) or shifting (bottom) the noise maps completely modifies the image. By contrast, our edit friendly noise maps enable editing while preserving structure (right).
}
\label{fig:generated_vs_us}
\end{figure}

Here we address the problem of inverting the DDPM scheme. As opposed to DDIM, in DDPM, $T+1$ noise maps are involved in the generation process, each of which has the same dimension as the generated output. Therefore, the total dimension of the noise space is larger than that of the output and there exist infinitely many noise sequences that perfectly reconstruct the image. While this property may provide flexibility in the inversion process, not every consistent inversion (\ie one that leads to perfect reconstruction) is also edit friendly. For example, one property we want from an inversion in the context of text-conditional models, is that fixing the noise maps and changing the text-prompt would lead to an artifact-free image, where the semantics correspond to the new text but the structure remains similar to that of the input image. What consistent inversions satisfy this property? A tempting answer is that the noise maps should be statistically independent and have a standard normal distribution, like in regular sampling. Such an approach was pursued in \cite{Wu22}. However, as we illustrate in Fig.~\ref{fig:generated_vs_us}, this native DDPM noise space is in fact not edit friendly. 

Here we present an alternative inversion method, which constitutes a better fit for editing applications, from text-guidance manipulations, to editing via hand-drawn colored strokes. Our inversion ``imprints'' the image more strongly onto the noise maps, which leads to better preservation of structure when fixing them and changing the condition of the model.
This is achieved by the fact that our noise maps have higher variances than the native ones. 
Our inversion requires no optimization and is extremely fast. Yet, it allows achieving state-of-the-art results on text-guided editing tasks with a relatively small number of diffusion steps, simply by fixing the noise maps and changing the text condition (\ie without requiring model fine-tuning or intervention in the attention maps). Importantly, our DDPM inversion can also be readily integrated with existing diffusion based editing methods that currently rely on approximate DDIM inversion. As we illustrate in Fig.~\ref{fig:teaser}, this improves their ability to preserve fidelity to the original image. Furthermore, since we find the noise vectors in a stochastic manner, we can provide a diverse set of edited images that all conform to the text prompt, a property not naturally available with DDIM inversion, see top row of Fig.~\ref{fig:teaser} and the Supplementary Supplementary (SM).

\section{Related work}
\label{related}



\subsection{Inversion of diffusion models}
Editing a real image using diffusion models requires extracting the noise vectors that would generate that image when used within the generative process. 
The vast majority of diffusion-based editing works use the DDIM scheme, which is a deterministic mapping from a single noise map to a generated image~\cite{Hertz22,Mokady22,Narek22,Guillaume22,Bram22,Parmar23,Adham22}. The original DDIM paper~\cite{Song21} suggested an efficient approximate inversion for that scheme. This method incurs a small error at every diffusion timestep, and these errors often accumulate into meaningful deviations when using classifier-free guidance~\cite{Ho21}. Mokady \etal~\cite{Mokady22} improve the reconstruction quality by fixing each timestep drifting. Their two-step approach first uses DDIM inversion to compute a sequence of noise vectors, and then uses this sequence to optimize the input null-text embedding at every timestep. Miyake \etal~\cite{miyake23}
achieve similar reconstruction accuracy through forward propagation without optimization, thereby enabling much faster editing processes. An improvement in the reconstruction quality was suggested by Han \etal~\cite{Han23} that integrate a regularization term into the null-text embedding optimization. 
EDICT~\cite{Bram22} enables mathematically exact DDIM-inversion of real images by maintaining two coupled noise vectors which are used to invert each other in an alternating fashion. This method doubles the computation time of the diffusion process. 
CycleDiffusion~\cite{Wu22} presents a DDPM-inversion method by recovering a sequence of noise vectors that perfectly reconstruct the image within the DDPM sampling process. As opposed to our method, their extracted noise maps are distributed like the native noise space of DDPM, which results in limited editing capabilities (see Figs.~\ref{fig:generated_vs_us},\ref{fig:cyclediffusion_vs_us}).



\subsection{Image editing using diffusion models}
The DDPM sampling method is not popular for editing of real images. When used, it is typically done without exact inversion. 
Two examples are Ho \etal~\cite{Ho20}, who interpolate between real images, and Meng \etal~\cite{Meng22} who edit real images via user sketches or strokes (SDEdit). Both construct a noisy version of the real image and apply a backward diffusion after editing. They suffer from an inherent tradeoff between the realism of the generated image and its faithfulness to the original contents. 
DiffuseIT~\cite{Gihyun22} performs image translation guided by a reference image or by text, also without explicit inversion. They guide the generation process by losses that measure similarity to the original image~\cite{Prafulla21}.

A series of papers apply text-driven image-to-image translation using DDIM inversion. Narek \etal~\cite{Narek22} and Cao \etal~\cite{cao23} achieve this by manipulating spatial features and their self-attention inside the model during the diffusion process. Hertz \etal~\cite{Hertz22} change the attention maps of the original image according to the target text prompt and inject them into the diffusion process. DiffEdit~\cite{Guillaume22} automatically generates a mask for the regions of the image that need to be edited, based on source and target text prompts. This is used to enforce the faithfulness of the unedited regions to the original image, in order to battle the poor reconstruction quality obtained from the inversion. This method fails to predict accurate masks for complex prompts.

Some methods utilize model optimization based on the target text prompt.  DiffusionCLIP \cite{Kim22} uses model fine-tuning based on a CLIP loss with a target text. Imagic \cite{Bahjat22} first optimizes the target text embedding, and then optimizes the model to reconstruct the image with the optimized text embedding. 
UniTune~\cite{Valevski22} also uses fine-tuning and shows great success in making global stylistic changes and complex local edits while maintaining image structure. Other works like Palette~\cite{Saharia21} and InstructPix2Pix~\cite{Brooks22}, learn conditional diffusion models tailored for specific editing tasks. 

\begin{figure}
\includegraphics[width=\columnwidth]{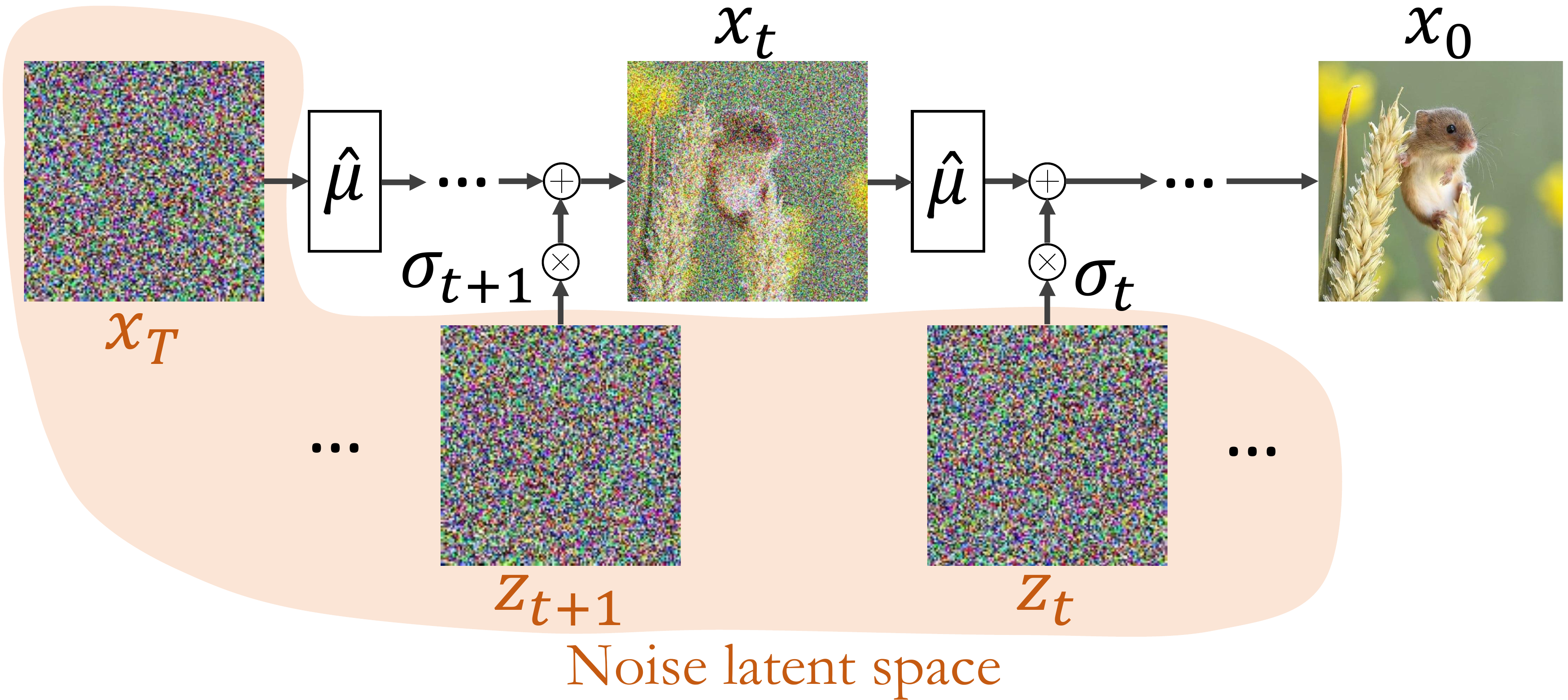}
\caption{\textbf{The DDPM latent noise space.} In DDPM, the generative (reverse) diffusion process synthesizes an image $x_0$ in $T$ steps, by utilizing $T+1$ noise maps, $\{x_T,z_T,\ldots,z_1\}$. We regard those noise maps as the latent code associated with the generated image.}
\label{fig:unfold}
\end{figure}







\section{The DDPM noise space}
\label{method}




Here we focus on the DDPM sampling scheme, which is applicable in both pixel space~\cite{Ho20} and latent space~\cite{Rombach22}. 
DDPM draws samples by attempting to reverse a diffusion process that gradually turns a clean image $x_0$ into white Gaussian noise,
\begin{equation}
\label{eq:xt_from_xtm1}
x_t = \sqrt{1-\beta_t} x_{t-1} + \sqrt{\beta_t}\, n_t,\quad t=1,\ldots,T,
\end{equation}
where $\{n_t\}$ are iid standard normal vectors and $\{\beta_t\}$ is some variance schedule. The forward process (\ref{eq:xt_from_xtm1}) can be equivalently expressed as
\begin{equation}
\label{eq:xt_from_x0}
x_t = \sqrt{\bar\alpha_t} x_0 + \sqrt{1-\bar\alpha_t}\, \epsilon_t,
\end{equation}
where $\alpha_t=1-\beta_t$, $\bar{\alpha}_t=\prod_{s=1}^t\alpha_s$, and $\epsilon_t\sim\mathcal{N}(0,\boldsymbol{I})$. It is important to note that in the representation \eqref{eq:xt_from_x0}, the vectors $\{\epsilon_t\}$ are \emph{not independent}. This is because each $\epsilon_t$ corresponds to the accumulation of the noises $n_1,\ldots,n_t$, so that $\epsilon_t$ and $\epsilon_{t-1}$ are highly correlated for all~$t$. This fact is irrelevant for the training process, 
which is not affected by the joint distribution of $\epsilon_t$'s across different timesteps, but it is important for our discussion below.

The generative (reverse) diffusion process starts from a random noise vector $x_T \sim \mathcal{N}(0,\mathbf{I})$ and iteratively denoises it using the recursion 
\begin{equation}
\label{eq:diffusion}
    x_{t-1} = \hat\mu_t(x_t) + \sigma_t z_t,\quad t=T,\ldots,1,
\end{equation}
where $\{z_t\}$ are iid standard normal vectors, and
\begin{align}
\label{eq:PD}
\hat\mu_t(x_t) = 
\sqrt{\bar \alpha_{t-1}} P(f_t(x_t)) + D(f_t(x_t)).
\end{align}
Here, $f_t$ is a neural network that is trained to predict  $\epsilon_t$ from $x_t$, $P(f_t(x_t))=(x_t-\sqrt{1-\bar \alpha_{t}}f_t(x_t))/\sqrt{\bar \alpha_{t}}$ is the predicted $x_0$, and $D(f_t(x_t))=\sqrt{1-\bar \alpha_{t-1}-\sigma_t^2} f_t(x_t)$ is a direction pointing to $x_t$. 
The variance schedule is taken to be $\sigma_t=\eta \beta_t (1-\bar{\alpha}_{t-1})/(1-\bar{\alpha}_t)$, where $\eta\in[0,1]$.
The case $\eta=1$ corresponds to the original DDPM work, and $\eta= 0$ corresponds to the deterministic DDIM scheme.

This generative process can be conditioned on text~\cite{Rombach22} or class~\cite{Ho21} by using a neural network $f$ that has been trained conditioned on those inputs. Alternatively, conditioning can be achieved through guided diffusion~\cite{Prafulla21,avrahami22}, which requires utilizing a pre-trained classifier or CLIP model during the generative process.


The vectors $\{x_T,z_T,\ldots,z_1\}$ uniquely determine the image $x_0$ generated by the process \eqref{eq:diffusion} (but not vice versa). We therefore regard them as the latent code of the model (see Fig.~\ref{fig:unfold}). Here, we are interested in the inverse direction. Namely, given a real image $x_0$, we would like to extract noise vectors that, if used in \eqref{eq:diffusion}, would generate $x_0$. We refer to such noise vectors as consistent with $x_0$. Our method, explained next, works with any $\eta\in(0,1]$.

\subsection{Edit friendly inversion}

It is instructive to note that \emph{any} sequence of $T+1$ images $x_0,\ldots,x_T$, in which $x_0$ is the real image, can be used to extract consistent noise maps by isolating $z_t$ from \eqref{eq:diffusion} as\footnote{Commonly, $z_1=0$ in DDPM, so that we run only over $t=T,\ldots,2$.} 
\begin{equation}
\label{eq:z_t}
z_t = \dfrac{x_{t-1} -\hat\mu_t(x_t)}{\sigma_t},\quad t=T,\ldots,1.
\end{equation}
However, unless such an auxiliary sequence of images is carefully constructed, they are likely to be far from the distribution of inputs on which the network $f_t(\cdot)$ was trained. In that case, fixing the so-extracted noise maps, $\{x_T,z_T,\ldots,z_1\}$, and changing the text condition, may lead to poor results.

What is a good way of constructing auxiliary images $x_1,\ldots,x_T$ for \eqref{eq:z_t} then? A naive approach is to draw them from a distribution that is similar to that underlying the generative process. Such an approach was pursued by~\cite{Wu22}. Specifically, they start by sampling  $x_T \sim \mathcal{N}(0,\mathbf{I})$. Then, for each $t=T,\ldots,1$ they isolate $\epsilon_t$ from \eqref{eq:xt_from_x0} using $x_t$ and the real $x_0$, substitute this $\epsilon_t$ for $f_t(x_t)$ in \eqref{eq:PD} to compute $\hat{\mu}_t(x_t)$, and use this $\hat{\mu}_t(x_t)$ in \eqref{eq:diffusion} to obtain $x_{t-1}$.

\begin{figure}
\centering
\includegraphics[width=0.94\columnwidth]{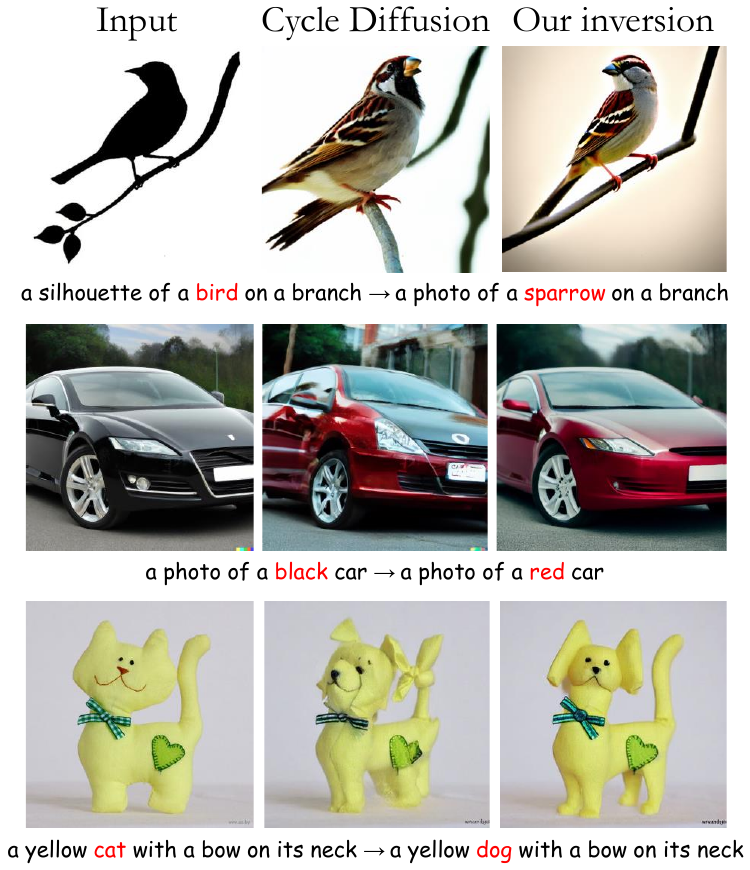}
\caption{\textbf{DDPM inversion via CycleDiffusion vs.~our method.} CycleDiffusion's inversion~\cite{Wu22} extracts a sequence of noise maps $\{x_T,z_T,\ldots,z_1\}$ whose joint distribution is close to that used in regular sampling. However, fixing this latent code and replacing the text prompt fails to preserve the image structure. Our inversion deviates from the regular sampling distribution, but better encodes the image structure.}
\label{fig:cyclediffusion_vs_us}
\end{figure}

The noise maps extracted by this method are distributed similarly to those of the generative process. Unfortunately, they are not well suited for editing global structures. This is illustrated in Fig.~\ref{fig:cyclediffusion_vs_us} in the context of text guidance and in Fig.~\ref{fig:shifting_bird} in the context of shifts. The reason for this is that DDPM's native noise space is not edit-friendly in the first place. Namely, even if we take a model-generated image, for which we have the ``ground-truth'' noise maps, fixing them while changing the text prompt does not preserve the structure of the image (see Fig.~\ref{fig:generated_vs_us}).

Interestingly, here we observe that constructing the auxiliary sequence $x_1,\ldots,x_T$ directly from $x_0$, and not via  (\ref{eq:xt_from_xtm1}), causes the image $x_0$ to be more strongly ``imprinted'' into the noise maps extracted from \eqref{eq:z_t}. 
Specifically, we propose to construct them as
\begin{equation}
\label{eq:xt_from_x0_iid}
x_t = \sqrt{\bar\alpha_t} x_0 + \sqrt{1-\bar\alpha_t}\, \tilde{\epsilon}_t,\quad 1,\ldots,T,
\end{equation}
where $\tilde{\epsilon}_t\sim\mathcal{N}(0,\boldsymbol{I})$ are \emph{statistically independent}. Note that despite the superficial resemblance between \eqref{eq:xt_from_x0_iid} and \eqref{eq:xt_from_x0}, these equations describe fundamentally different stochastic processes. In \eqref{eq:xt_from_x0} every pair of consecutive $\epsilon_t$'s are highly correlated, while in \eqref{eq:xt_from_x0_iid} the $\tilde{\epsilon}_t$'s are independent. This implies that in our construction, $x_t$ and $x_{t-1}$ are typically farther away from each other than in \eqref{eq:xt_from_x0}, so that every $z_t$ extracted from \eqref{eq:z_t} has a higher variance than in the regular generative process. A pseudo-code of our method is provided in Alg.~\ref{alg:example}. 


A few comments are in place regarding this inversion method. First, it reconstructs the input image up to machine precision, given that we compensate for accumulation of numerical errors (last row in Alg.~\ref{alg:example}), as we explain in the SM. Second, it is straightforward to use with any kind of diffusion process (\eg a conditional model~\cite{Ho2021Cascaded,Ho20,Rombach22}, guided-diffusion~\cite{Prafulla21},
classifier-free~\cite{Ho21}) by using the appropriate form for $\hat\mu_t(\cdot)$. Lastly, due to the randomness in \eqref{eq:xt_from_x0_iid}, we can obtain many different inversions. While each of them leads to perfect reconstruction, when used for editing they will lead to different variants of the edited image. This allows generating diversity in \eg text-based editing tasks, a feature not naturally available with DDIM inversion methods (see Fig.~\ref{fig:teaser} and SM).

\begin{algorithm}[tb]
   \caption{Edit-friendly DDPM inversion}
   \label{alg:example}
\begin{algorithmic}
   \STATE {\bfseries Input:} real image $x_0$ 
   \STATE {\bfseries Output:} $\{x_T,z_T,\ldots,z_1\}$
   \FOR{$t=1$ {\bfseries to} $T$}
   \STATE $\tilde{\epsilon} \sim \mathcal{N}(0,\,1)$
   \STATE $x_t \leftarrow \sqrt{\bar{\alpha_t}} x_0 + \sqrt{1-\bar{\alpha_t}} \tilde{\epsilon}$
   \ENDFOR
   \FOR{$t=T$ {\bfseries to} $1$}
   \STATE $z_t \leftarrow (x_{t-1} -\hat\mu_t(x_{t})) / \sigma_t $   
   \STATE $x_{t-1} \leftarrow \hat\mu_t(x_{t})+\sigma_t z_t$ \quad // to avoid error accumulation
   \ENDFOR
   \STATE {\bfseries Return:} $\{x_T,z_T,\ldots,z_1\}$
\end{algorithmic}
\end{algorithm}

\subsection{Properties of the edit-friendly noise space}
\label{sec:properties}

\begin{figure}
\begin{center}
\hspace{0.45cm} Regular dynamics \hspace{1.1cm} Edit friendly dynamics\\
\hspace{0.2cm}\includegraphics[width=0.47\columnwidth, trim={2.2cm 0cm 2.2cm 1.2cm},clip]{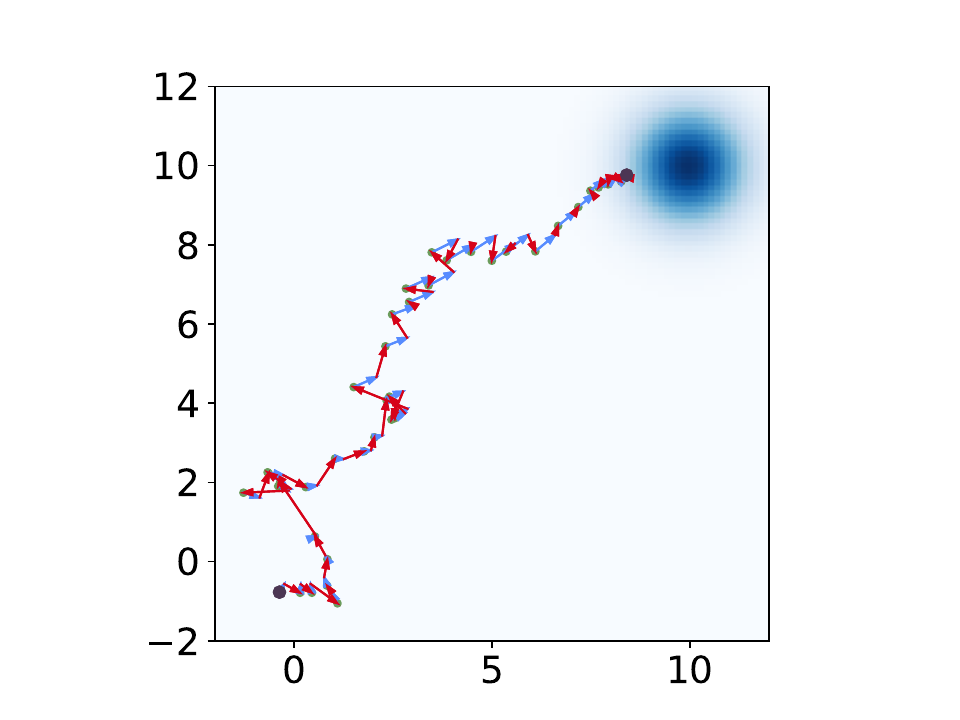}
\hspace{0.2cm}\includegraphics[width=0.47\columnwidth, trim={2.2cm 0cm 2.2cm 1.2cm},clip]{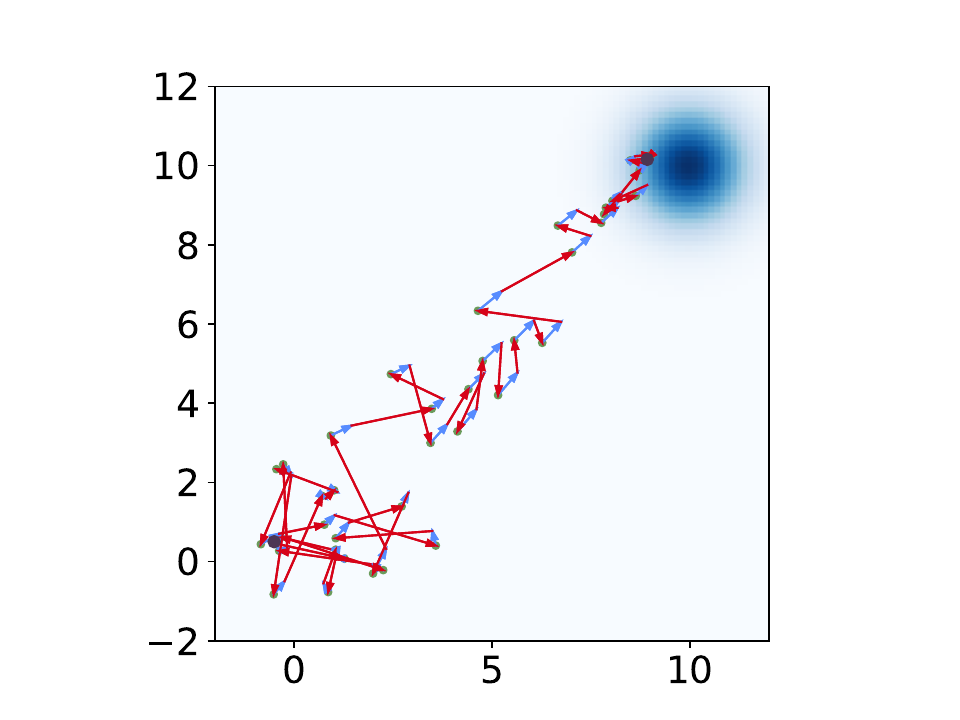}
\includegraphics[width=0.49\columnwidth, trim={0cm 0cm 0cm 0.5cm},clip]{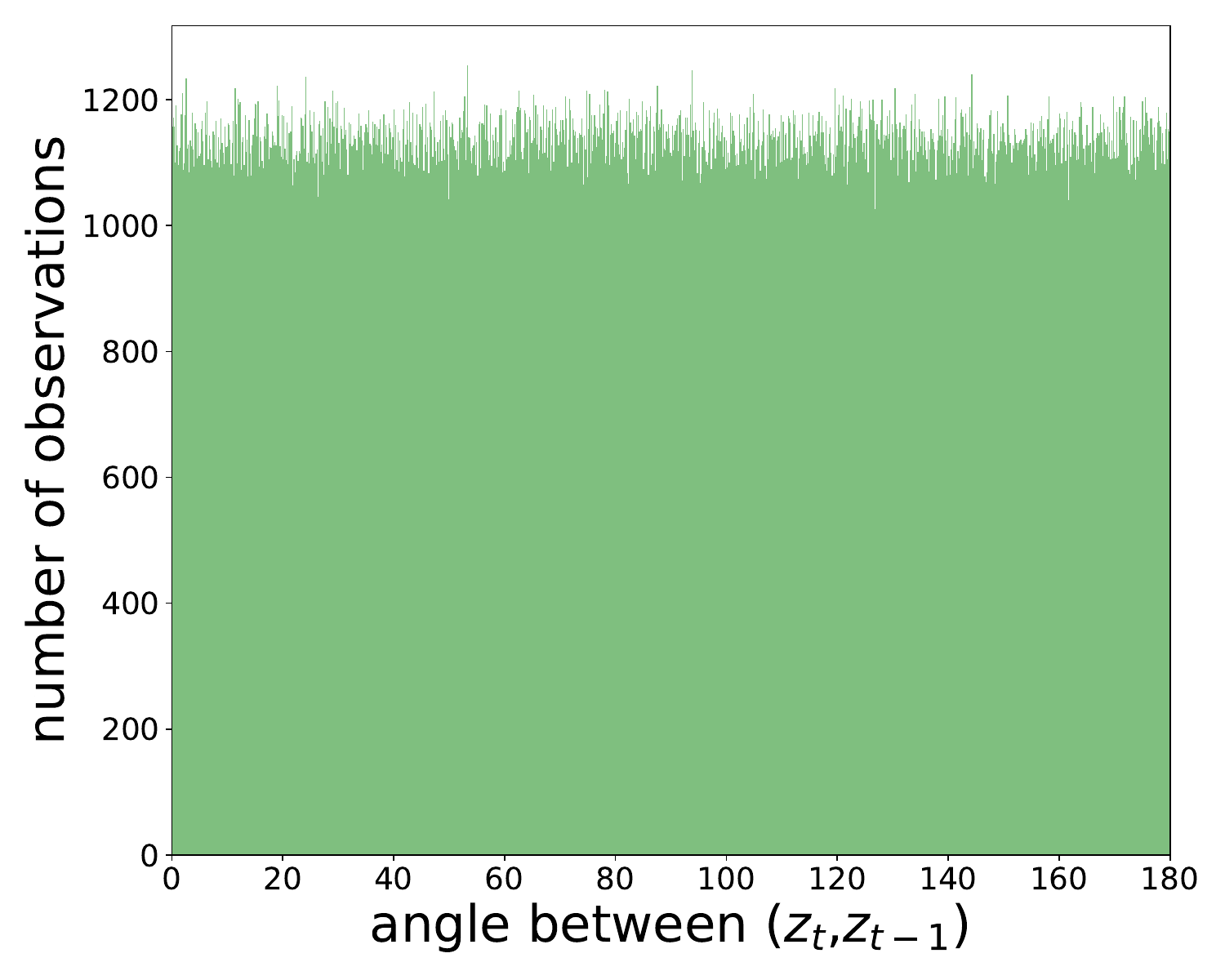}
\includegraphics[width=0.49\columnwidth, trim={0cm 0cm 0cm 0.5cm},clip]{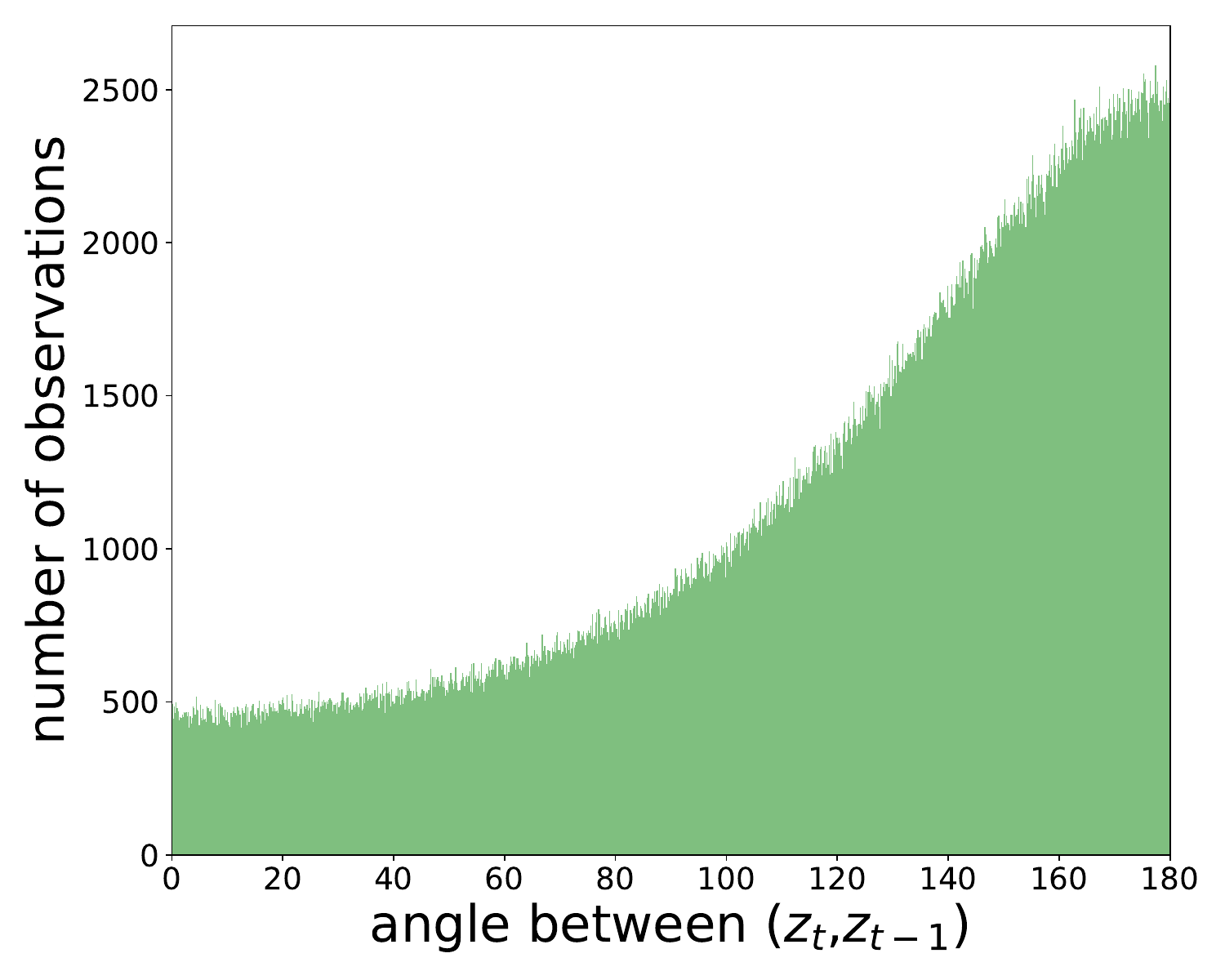}
\end{center}
\caption{\textbf{Regular vs.~edit-friendly diffusion.} In the regular generative process (top left), the noise vectors (red) are statistically independent across timesteps and thus the angle between consecutive vectors is uniformly distributed in $[0,180^\circ]$ (bottom left). In our dynamics (top right) the noise vectors have higher variances and are negatively correlated across consecutive times (bottom right).}
\label{fig:statistics_2d}
\end{figure}

    We now explore the properties of our edit-friendly noise space and compare it to the native DDPM noise space. We start with a 2D illustration, depicted in Fig.~\ref{fig:statistics_2d}. Here, we use a diffusion model designed to sample from $\mathcal{N}( \begin{psmallmatrix}10\\10\end{psmallmatrix},\mathbf{I})$. The top-left pane shows a regular DDPM process with $40$ inference steps. It starts from $x_T \sim \mathcal{N}(\begin{psmallmatrix}0\\0\end{psmallmatrix},\mathbf{I})$ (black dot at the bottom left), and generates a sequence $\{x_t\}$ (green dots) that ends in $x_0$ (black dot at the top right). Each step is broken down to the deterministic drift $\hat\mu_t (x_t)$ (blue arrow) and the noise vector $z_t$ (red arrow). On the top-right pane, we show a similar visualization, but for our latent space. Specifically, here we compute the sequences $\{x_t\}$ and $\{z_t\}$ using Alg.~\ref{alg:example} for some given $x_0\sim \mathcal{N}(\begin{psmallmatrix}10\\10\end{psmallmatrix},\mathbf{I})$. As can be seen, in our case, the noise perturbations $\{z_t\}$ are larger. This property comes from our construction of $x_t$, which are typically farther away from one another than in \eqref{eq:xt_from_x0}. How can the red arrows be longer and still form a trajectory from the origin to the blue cloud? Close inspection reveals that the angles between consecutive noise vectors tend to be obtuse. In other words, our noise vectors are (negatively) correlated across consecutive times. This can also be seen from the two plots in the bottom row, which depict the histograms of angles between consecutive noise vectors for the regular sampling process and for ours. In the former case, the angle distribution is uniform, and in the latter it has a peak at $180^\circ$. 

\begin{figure}
\includegraphics[scale=0.28, trim={0.8cm 0cm 1.2cm 1cm},clip]{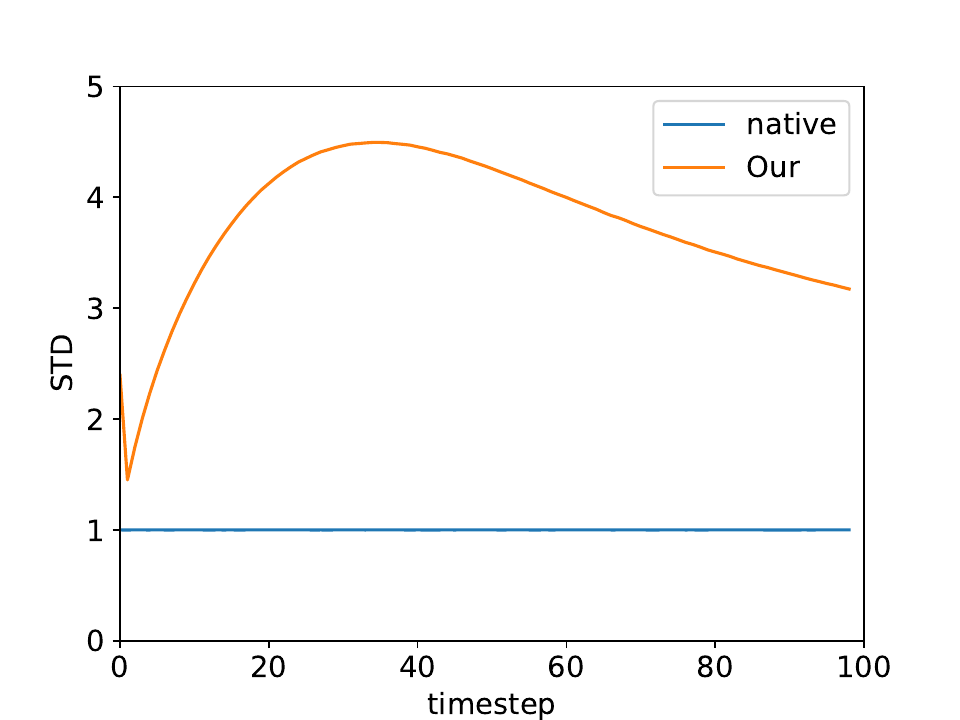}
\includegraphics[scale=0.28, trim={0cm 0cm 1.2cm 1.2cm},clip]{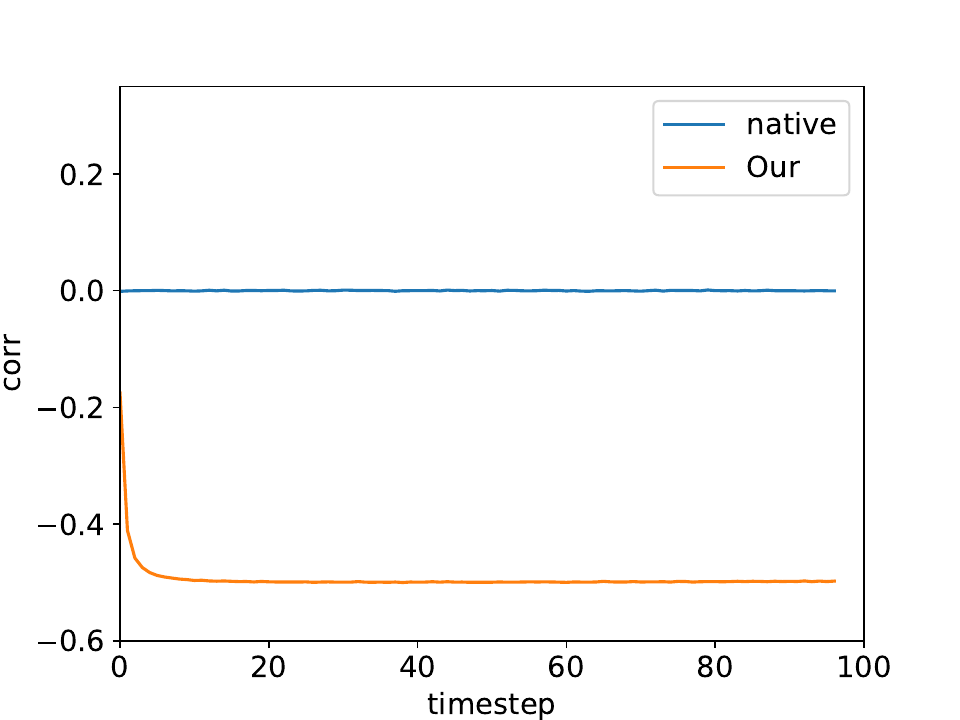}
\caption{\textbf{Native vs.~edit friendly noise statistics.} Here we show the per-pixel standard deviations of $\{z_t\}$ and the per-pixel correlation between them for model-generated images.}
\label{fig:statistics_images}
\end{figure}

The same qualitative behavior occurs in diffusion models for image generation. Figure~\ref{fig:statistics_images} shows the per-pixel variance of $z_t$ and correlation between $z_t$ and $z_{t-1}$ for sampling with 100 steps from an unconditional diffusion model trained on Imagenet. Here the statistics were calculated over 10 images drawn from the model. As in the 2D case, our noise vectors have higher variances and they exhibit negative correlations between consecutive steps. As we illustrate next, these larger variance noise vectors encode the structure of the input image more strongly, and are thus more suitable for editing. 


\begin{figure}
\centering
\includegraphics[width=\columnwidth]{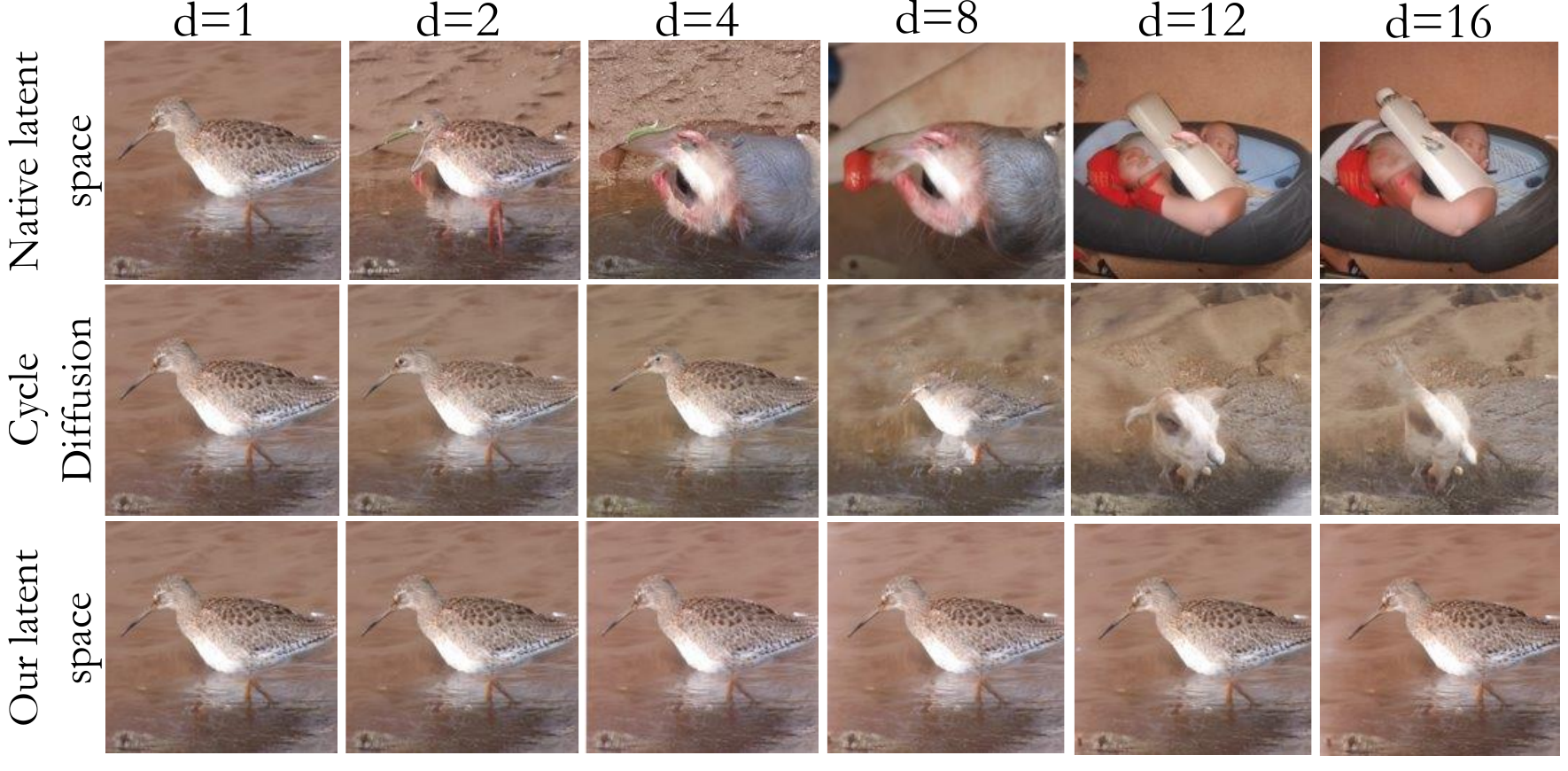}
\caption{\textbf{Image shifting.} We shift to the right a $256 \times 256$ image generated by an unconditional model trained on ImageNet by $d=1\ldots16$ pixels. When shifting the native noise maps (top) or the ones extracted by CycleDiffusion \cite{Wu22} (middle) the structure is lost. With our latent space, the structure is preserved.}
\label{fig:shifting_bird}
\end{figure}

\vspace{-0.4cm}
\paragraph{Image shifting}
Intuitively, shifting an image should be possible by shifting all $T+1$ maps of the latent code. 
Figure~\ref{fig:shifting_bird} shows the result of shifting the latent code of a model-generated image by various amounts. As can be seen, shifting the native latent code (the one used to generate the image) leads to a complete loss of image structure. In contrast, shifting our edit-friendly code, results in minor degradation. Quantitative evaluation is provided in the SM.



\begin{figure}
\centering
\vspace{-0.05in}
\hspace{1in}

\includegraphics[width=0.9\columnwidth]{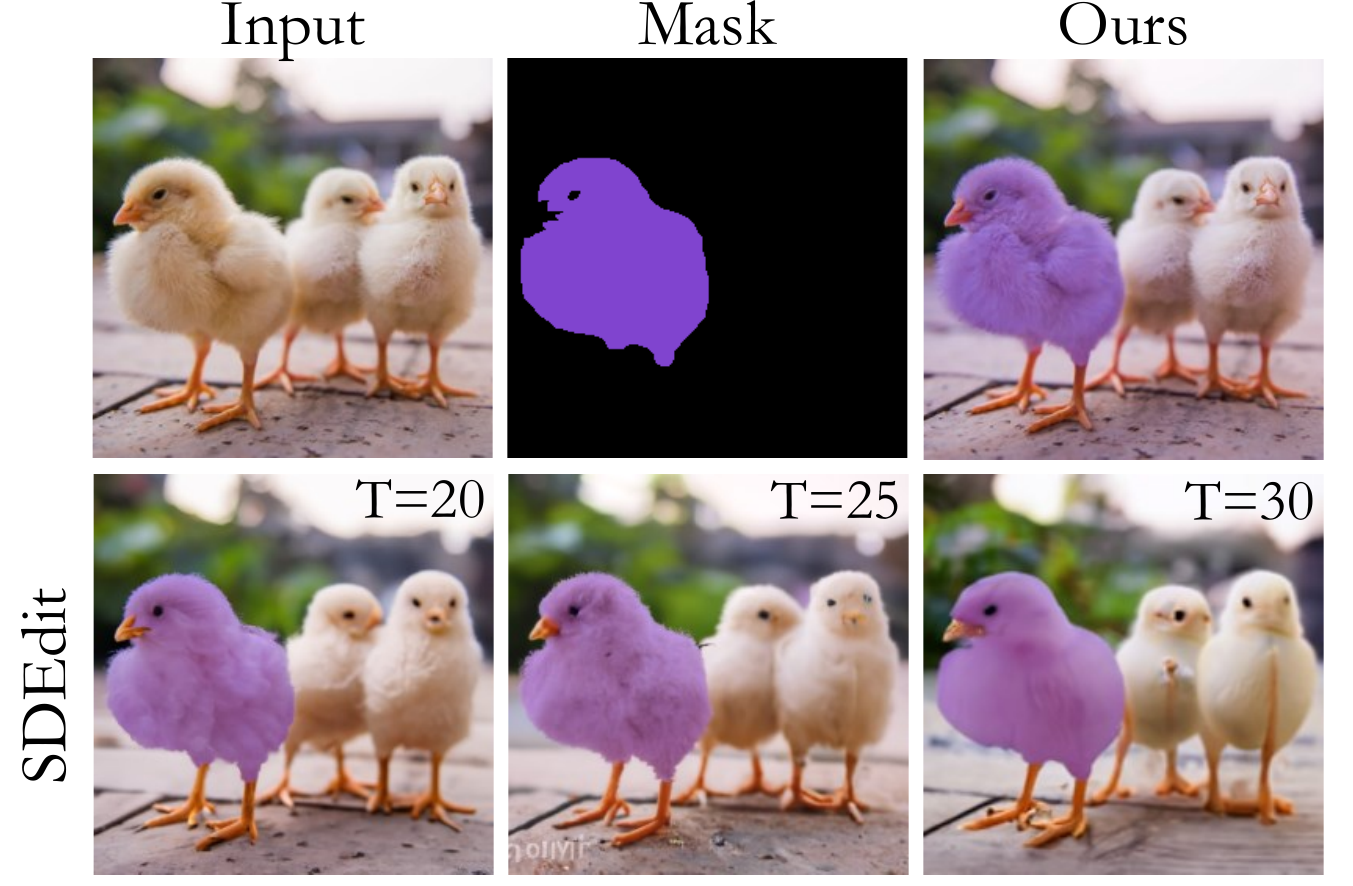}
\caption{\textbf{Color manipulation on a real image.} Our method (applied here from $T_2=70$ to $T_1=20$ with $s=0.05$) leads to a strong editing effect without modifying textures and structures. SDEdit, on the other hand, either does not integrate the mask well when using a small noise (left) or does not preserve structures when the noise is large (right). In both methods, we use an unconditional model trained on ImageNet with $100$ inference steps.} 
\label{fig:mask_color}
\end{figure}

\vspace{-0.4cm}
\paragraph{Color manipulations}
Our latent space also enables convenient manipulation of color. 
Specifically, suppose we are given an input image $x_0$, a binary mask $B$, and a corresponding colored mask $M$. We start by constructing $\{x_1,\ldots,x_T\}$ and extracting $\{z_1,\ldots,z_T\}$ using \eqref{eq:xt_from_x0_iid} and \eqref{eq:z_t}, as before. Then, we modify the noise maps as
\begin{equation}
\label{eq:zt_mask}
z_t^{\text{edited}} = z_t + s B\odot ( M-P(f_t(x_t))),
\end{equation}
with $P(f_t(x_t))$ from \eqref{eq:PD}, where $s$ is a parameter controlling the editing strength. We perform this modification over a range of timesteps $[T_1,T_2]$. Note that the term in parenthesis encodes the difference between the desired colors and the predicted clean image in each timestep. 
Figure~\ref{fig:mask_color} illustrates the effect of this process in comparison to SDEdit, which suffers from an inherent tradeoff between fidelity to the input image and conformation to the desired edit. Our approach can achieve a strong editing effect without modifying textures (neither inside nor outside the mask). 


\vspace{-0.1cm}
\section{Text-Guided Image Editing} 
\begin{figure*}
\centering
\includegraphics[width=0.8\textwidth]{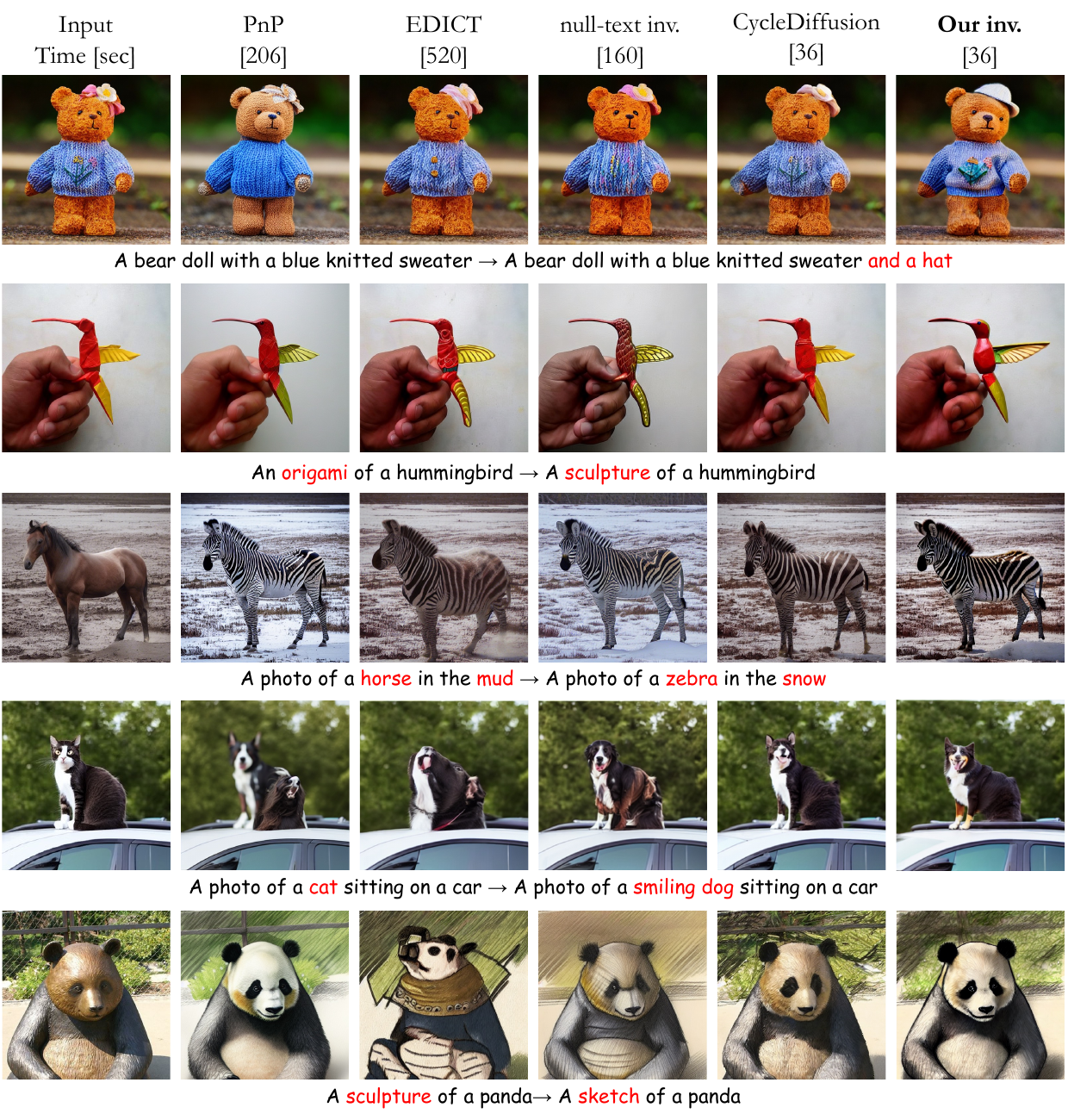}
\caption{\textbf{Comparisons.} We show results for editing of real images using all methods. Our approach maintains high fidelity to the input while conforming to the text prompt. The time taken to edit a single image is indicated within parentheses.}
\label{fig:comparisons}
\end{figure*}

Our latent space can be utilized for text-driven image editing. 
\label{sec:text-guided}
Suppose we are given a real image $x_0$, a text prompt describing it $p_{\text{src}}$, and a target text prompt $p_{\text{tar}}$. To modify the image according to these prompts, we extract the edit-friendly noise maps $\{x_T,z_T,\ldots,z_1\}$, while injecting $p_{\text{src}}$ to the denoiser. We then fix those noise maps and generate an image while injecting $p_{\text{tar}}$ to the denoiser. We run the generation process starting from timestep $T\!-\!T_{\text{skip}}$, where $T_{\text{skip}}$ is a parameter controlling the adherence to the input image.
Figures~\ref{fig:teaser},~\ref{fig:generated_vs_us},~\ref{fig:cyclediffusion_vs_us}, and \ref{fig:comparisons} show several text driven editing examples using this approach. As can be seen, this method nicely modifies semantics while preserving the structure of the image. In addition, it allows generating diverse outputs for any given edit (see Fig.~\ref{fig:teaser} and SM).
We further illustrate the effect of using our inversion in combination with methods that rely on DDIM inversion (Figs. ~\ref{fig:teaser} and \ref{fig:zero-short-qualitative}). As can be seen, these methods often do not preserve fine textures, like fur, flowers, or leaves of a tree, and oftentimes also do not preserve the global structure of the objects. By integrating our inversion, structures and textures are better preserved.


\section{Experiments}
\label{sec:evalution}

We evaluate our method both quantitatively and qualitatively on text-guided editing of real images. We analyze the usage of our extracted latent code by itself (as explained in Sec.~\ref{sec:text-guided}), and in combination with existing methods that currently use DDIM inversion. In the latter case, we extract the noise maps using our DDPM-inversion and inject them in the reverse process, in addition to any manipulation they perform on \eg attention maps. All experiments use a real input image as well as source and target text prompts.


\vspace{-0.3cm}
\paragraph{Implementation details} We use Stable Diffusion~\cite{Rombach22}, in which the diffusion process is applied in the latent space of a pre-trained image autoencoder. The image size is $512\! \times 512 \!\times \!3$, and the latent space is $64 \! \times 64 \! \times \! 4$. Our method is also applicable in unconditional pixel space models using CLIP guidance, however we found Stable Diffusion to lead to better results. 
Two hyper-parameters control the balance between faithfulness to the input image and adherence to the target prompt: the strength of the classifier-free guidance~\cite{Ho21}, and $T_{\text{skip}}$ explained in Sec.~\ref{sec:text-guided}. In all our numerical analyses and all the results in the paper, we used $\text{strength}\!=\!15$,  $T_{\text{skip}}\!=\!36$, $\eta\!=\!1$ and $100$ inference steps,  unless noted otherwise. 
In the SM we provide a thorough analysis of the effect of the strength and $T_{\text{skip}}$ hyper-parameters.

\begin{figure}[h]
\centering
\includegraphics[trim={0 0.7cm 0 2.1cm},clip=true,width=1\columnwidth]{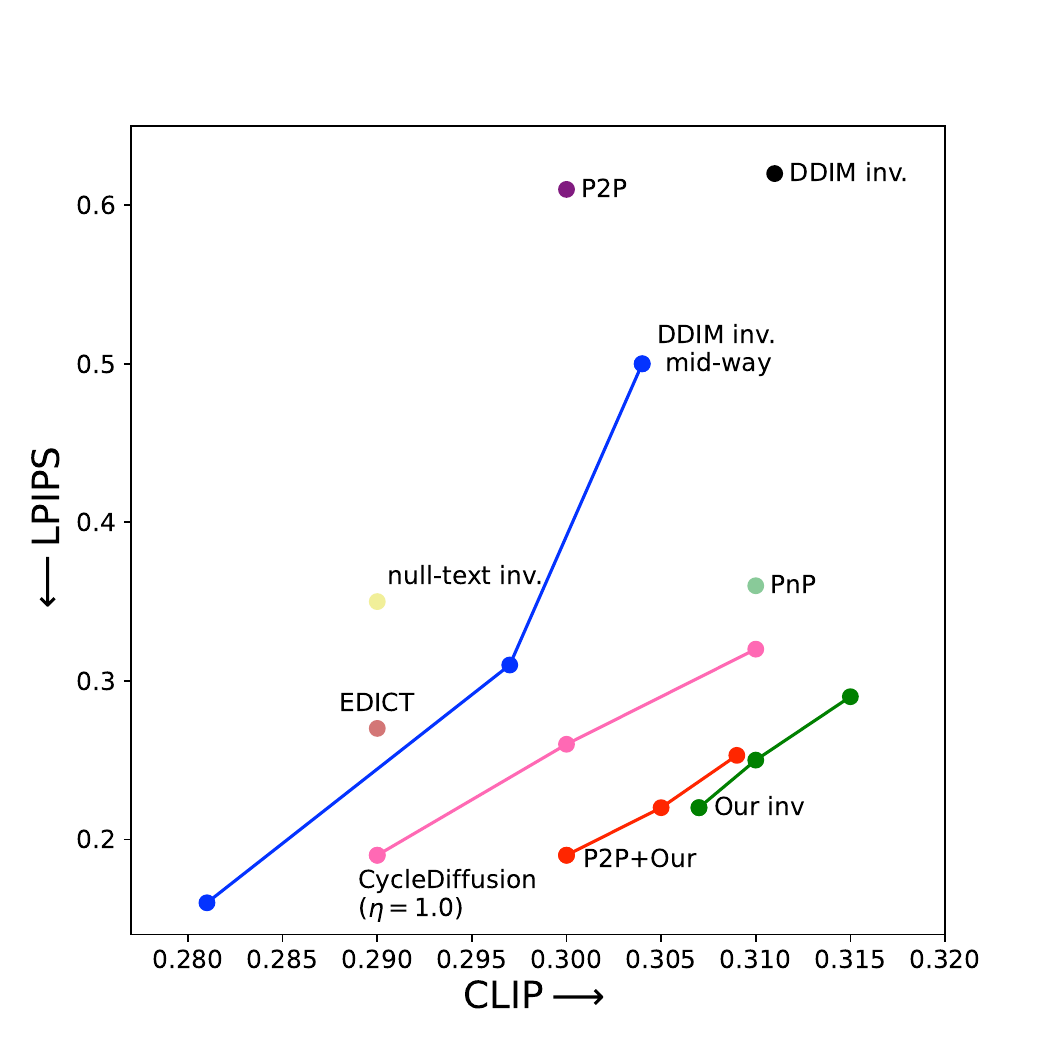}
\caption{\textbf{Fidelity to source image vs.~compliance with target text.} The plot compares the LPIPS and CLIP scores achieved by all methods on the modified ImageNet-R-TI2I dataset.  Our inversion, P2P with our inversion, and CycleDiffusion are shown with three options of the parameters (strength,$T_{\text{skip}}$): for our method $(15,36),(12,36),(9,36)$, for P2P+Ours $(7.5,8),(7.5,12),(9,20)$, and for CycleDiffusion $(3,30),(4,25),(4,15)$. DDIM inversion mid-way is shown with three options for $T_{\text{skip}}$: $20, 40, 60$, all with a guidance strength of 9. The parameters for the other methods are reported in the SM. For CLIP higher is better, while for LPIPS lower is better.}
\label{fig:clip_lpips}
\end{figure}


\vspace{-0.4cm}
\paragraph{Datasets} We use two datasets of real images: (i)~``modified ImageNet-R-TI2I'' from~\cite{Narek22} with additional examples collected from the internet and from other datasets, (ii)~``modified Zero-Shot I2IT'', which contains images of 4 classes (Cat, Dog, Horse, Zebra) from~\cite{Parmar23} and from the internet. The first dataset comprises 48 images, with 3-5 different target prompts for each, leading to 212 image-text pairs. The second dataset has 15 images in each category with one target prompt for each, making 60 image-text pairs in total. Please refer to the SM for full details.

\vspace{-0.4cm}
\paragraph{Metrics} We numerically evaluate the results using two complementary metrics:  LPIPS~\cite{zhang18} to quantify the extent of structure preservation (lower is better) and a CLIP-based score to quantify how well the generated images comply with the text prompt (higher is better). We additionally quantify the editing time in seconds required for image editing. Further information on diversity is provided in the SM.

\vspace{-0.4cm}
\paragraph{Comparisons on the modified ImageNet-R-TI2I dataset} 
We perform comparisons with Plug-and-Play (PnP)~\cite{Narek22}, EDICT~\cite{Bram22}, null-text inversion~\cite{Mokady22} and CycleDiffusion~\cite{Wu22}. Additionally, we assess our method against prompt-to-prompt (p2p) \cite{Hertz22}, both as a standalone technique and when integrated with our inversion. Further details on the integration can be found in the SM. We report the results of CycleDiffusion with $\eta=1$, similarly to our configuration. Quantitative results with $\eta=0.1$, as suggested in their paper, can be found in SM. Finally, we compare our method to both plain DDIM inversion and DDIM inversion that applies the inversion until a specific timestep. All methods were run with the default parameters suggested by their authors and are provided in SM. For a fair comparison, we use the same parameters of CycleDiffusion across all images within the experiment. This is in contrast to their paper where parameters are individually selected for each image. 

As seen in Fig.~\ref{fig:comparisons}, our method successfully modifies real images according to the target prompts. In all cases, our results exhibit both high fidelity to the input image and adherence to the target prompt. EDICT shows some artifacts in their results and CycleDiffusion produces images with less compliance with the target text. PnP and null-text inversion often preserve structure but require more than 2.5 minutes to edit an image. Qualitative results with plain DDIM inversion and P2P with and without our inversion appear in SM. Figure~\ref{fig:clip_lpips} shows the CLIP-LPIPS losses graph for all methods, where, for our inversion, P2P with our inversion, and CycleDiffusion we report these losses with three different parameters. As can be seen, our method achieves a good balance between LPIPS and CLIP. CycleDiffusion struggles to apply a strong edit while preserving the structure. Integrating our inversion into P2P improves their performance in both metrics. See more details in the SM.

\begin{figure}
\centering
\includegraphics[width=0.96\columnwidth]{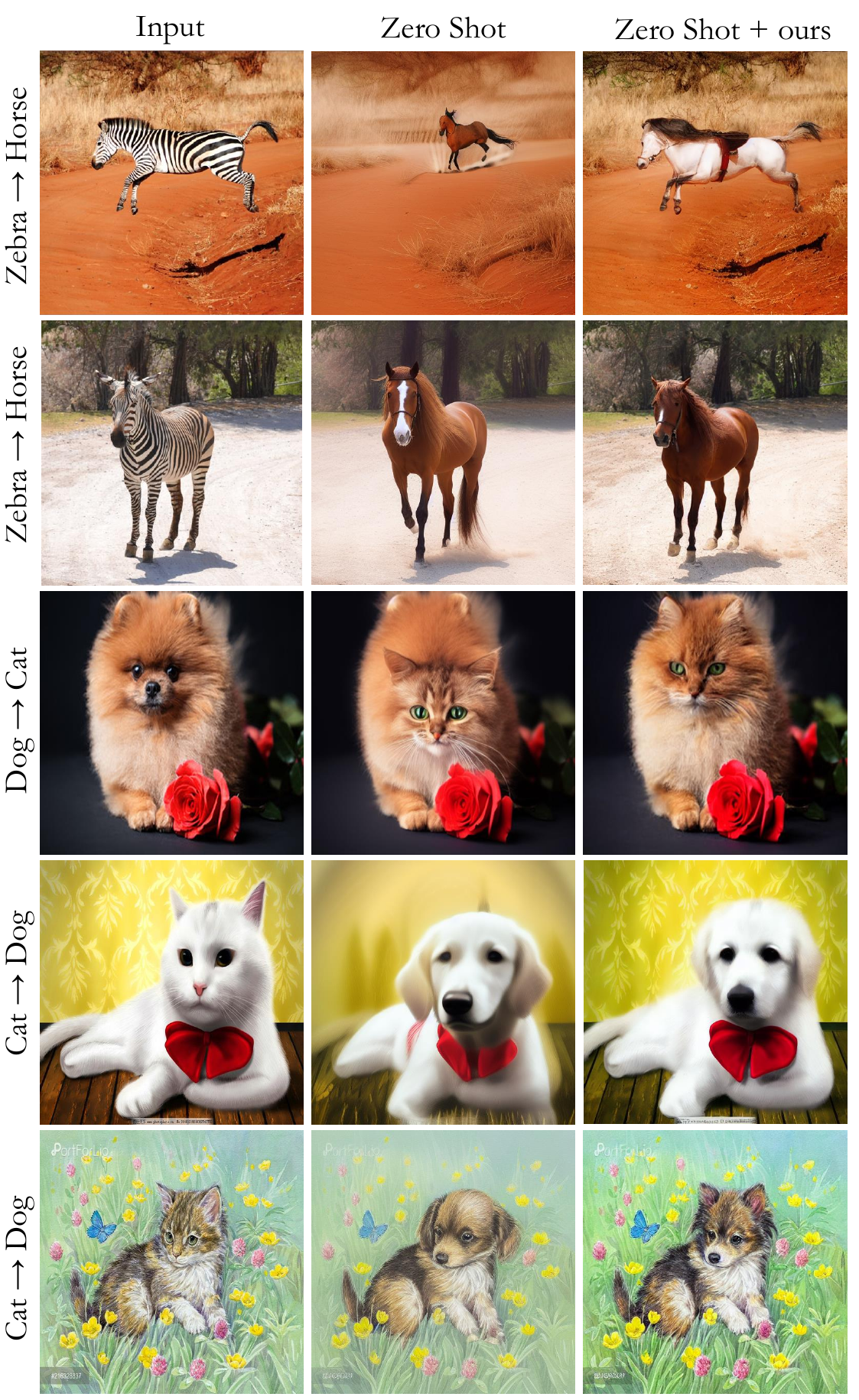}
\caption{\textbf{Improving Zero-Shot I2I Translation.} Images generated by Zero Shot I2I suffer from loss of detail. With our inversion, fine textures, like fur and flowers, are retained. 
Both methods achieve CLIP accuracy of 0.88, however, Zero-Shot method achieves LPIPS score of 0.35 while Zero-Shot with our inversion produces images more similar to the input, and hence, achievs an LPIPS score of 0.27 (for LPIPS lower is better).}
\label{fig:zero-short-qualitative}
\end{figure}


\vspace{-0.3cm}
\paragraph{Comparisons on the modified Zero-Shot I2IT dataset}
Next, we compare our method to Zero-Shot Image-to-Image Translation (Zero-Shot)~\cite{Parmar23}, which uses DDIM inversion.
This method only translates between several predefined classes. We follow their setting and use $50$ diffusion steps. When using with our inversion, we decrease the hyper-parameter controlling the cross-attention from the default value $0.1$ to $0.03$. As can be seen in Fig.~\ref{fig:zero-short-qualitative}, while Zero-Shot's results comply with the target text, they are typically blurry and miss detail. Integrating our inversion adds back the details from the input image. See more details in the SM.

\section{Conclusion}
\label{conclusion}

We presented an inversion method for DDPM. Our noise maps encode the image structure more strongly than the noise maps in regular sampling, and are therefore better suited for image editing. We illustrated their advantages in text-based editing, both when used by themselves and in combination with other editing methods. 





{
    \small
    \bibliographystyle{ieeenat_fullname}
    \bibliography{main}
}

\clearpage
\onecolumn
\appendix


\renewcommand\thefigure{S\arabic{figure}}    
\setcounter{figure}{0}  
\renewcommand\theequation{S\arabic{equation}}    
\setcounter{equation}{0}
\renewcommand\thealgorithm{S\arabic{algorithm}}    
\setcounter{algorithm}{0}  
\renewcommand{\thesection}{\Alph{section}}
\renewcommand\thetable{S\arabic{table}}

\section{Shifting the latent code}\label{app:shift}
\begin{figure}[H]
\centering
\includegraphics[width=0.5\textwidth]{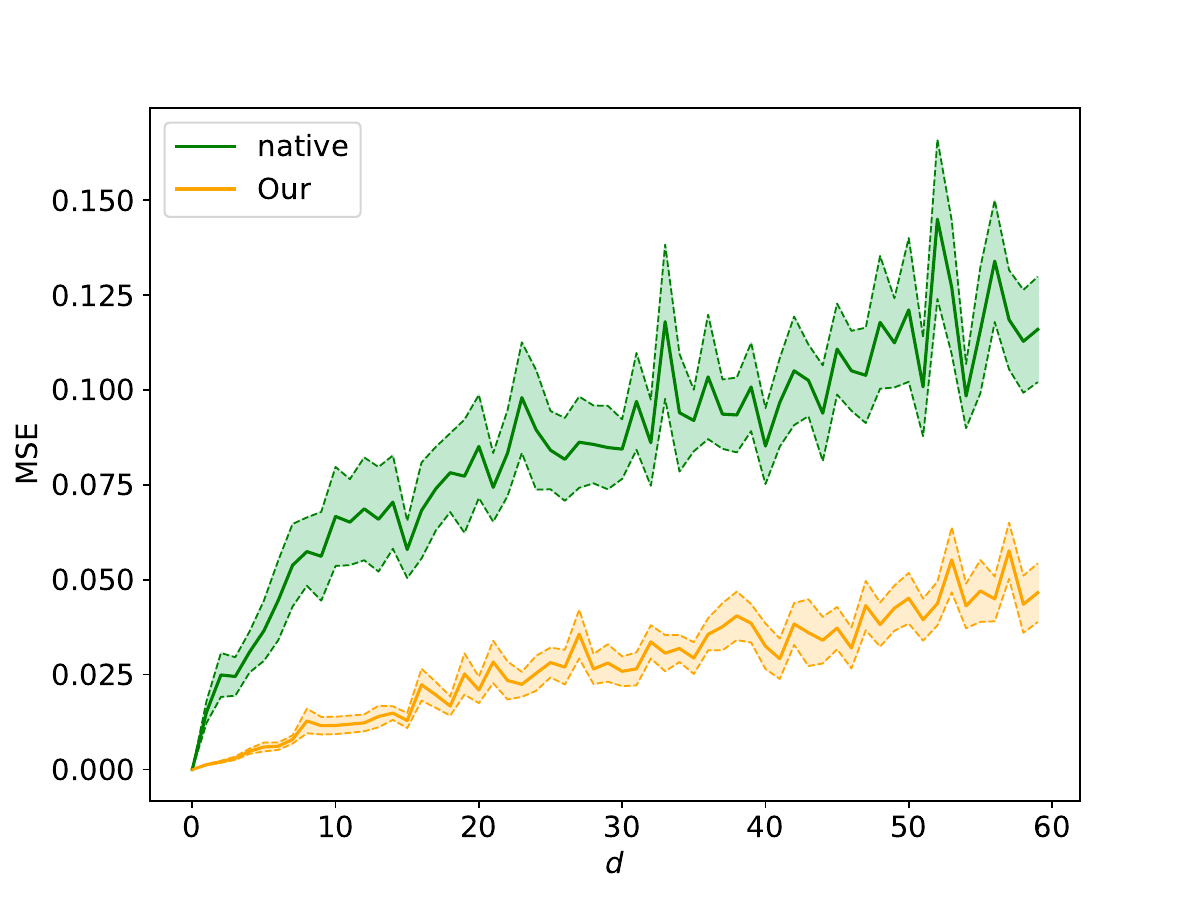}
\caption{\textbf{Shifting the latent code}. We plot the MSE over the valid pixels after shifting the latent code and generating the image. The colored regions represent one standard error of the mean (SEM) in each direction.}
\label{fig:shifting_graph}
\end{figure}

As described in Sec.~\ref{method}, we can shift an input image by shifting its extracted latent code. This requires inserting new columns/rows at the boundary of the noise maps. To guarantee that the inserted columns/rows are drawn from the same distribution as the rest of the noise map, we simply copy a contiguous chunk of columns/rows from a different part of the noise map. In all our experiments, we copied into the boundary the columns/rows indexed $\{50,\ldots,50+d-1\}$ for a shift of $d$ pixels. We found this strategy to work better than randomly drawing the missing columns/rows from a white normal distribution having the same mean and variance as the rest of the noise map. 
Figure~\ref{fig:shifting_graph} depicts the MSE over the valid pixels that is incurred when shifting the noise maps. This analysis was done using 25 model-generated images. As can be seen, shifting our edit-friendly code results in minor degradation while shifting the native latent code leads to a complete loss of the image structure. 

\clearpage

\section{The effect of the numerical error} 
\begin{figure}[H]
\centering
\includegraphics[width=\columnwidth]{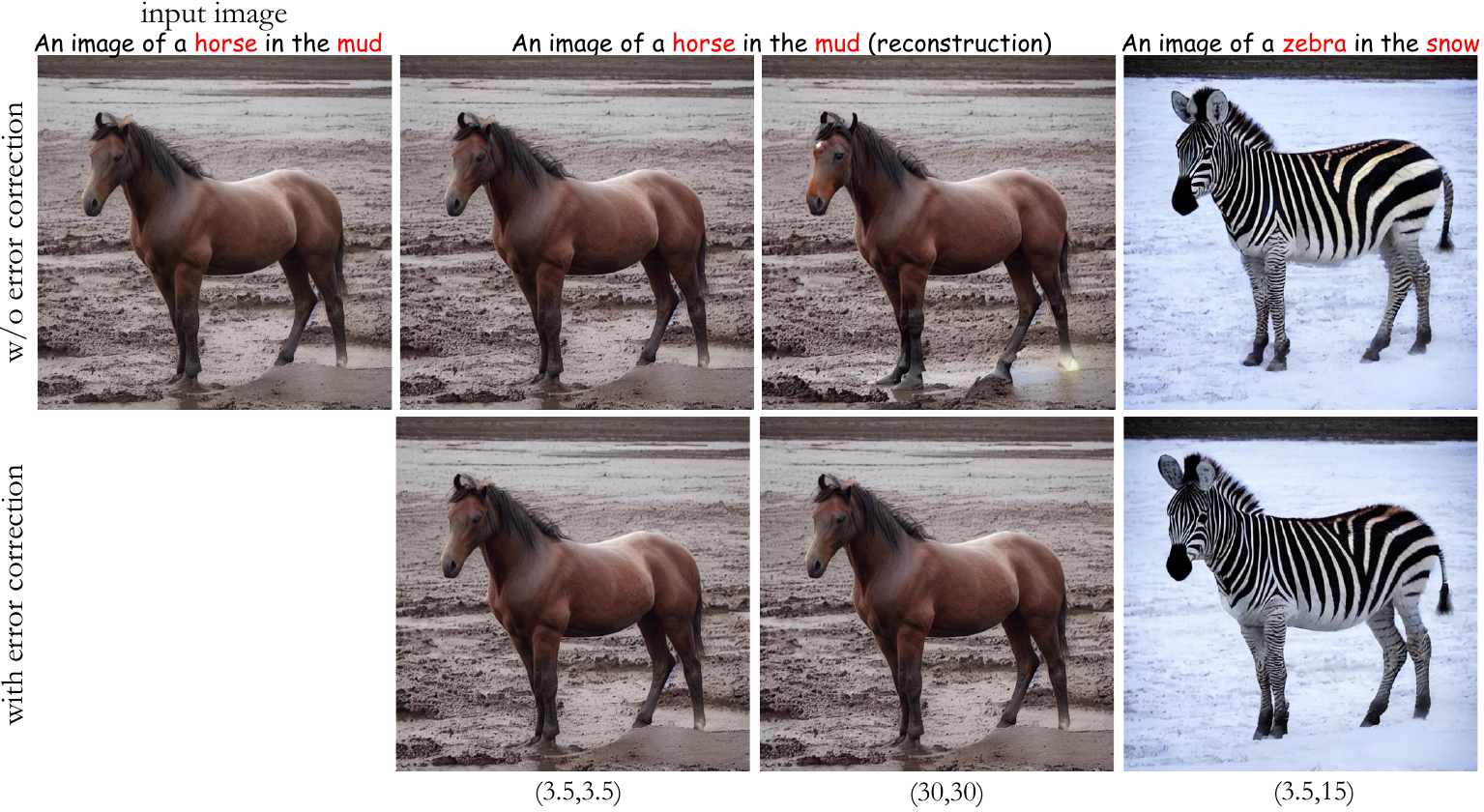}
\caption{\textbf{Error correction effect.} 
Below the images, we specify the strength parameters used for the inversion (first number within the parentheses) and the sampling (second number within the parentheses). Above the images, we specify the prompt used. Above the leftmost column is $p_{\text{src}}$, while above the other columns is $p_{\text{tar}}$. The parameter $T_{\text{skip}}$ is set to 36, as in our experiments in the main text. The second and third columns show reconstructions. As can be seen, with a large strength parameter, the reconstruction is not perfect without the correction (\eg the head and the leg of the horse). However, this numerical drifting does not influence the editing quality (rightmost column).}

\label{fig:reconstruction_for_rebuttal}
\end{figure}

In Algorithm~\ref{alg:example} we add a correction step for avoiding numerical drifting. This step assists in achieving perfect reconstruction. Note that in order to reconstruct the input image, the hyper-parameters used to extract the noise maps should be identical to the ones used for sampling. Specifically, the prompt, $T_{\text{skip}}$, and strength parameters should be the same in the function $\mu_t(x_t)$ used for the inversion (Eq.~\ref{eq:z_t}) and in the function $\mu_t(x_t)$ used during sampling (Eq.~\ref{eq:diffusion}). We note that the effect of the numerical drifting is noticeable only when using a large strength parameter (see second and third column in Fig.~\ref{fig:reconstruction_for_rebuttal}). By default, when performing text-based editing, we do not use extreme values for the strength parameter, and therefore in such cases this correction is not needed (rightmost column in Fig.~\ref{fig:reconstruction_for_rebuttal}).

We calculate the PSNR between the images with and without the correction for the example that appears in Fig.~\ref{fig:reconstruction_for_rebuttal}. In the reconstruction case, using $\text{strength}=30$, the PSNR can drop to below 17dB. As noted, this correction is not needed for editing, where the PSNR between the edited images with and without the correction is 67.4dB.

\clearpage

\section{CycleDiffusion}
\begin{figure}[H]
\includegraphics[width=0.485\textwidth, trim={0.8cm 0cm 1.2cm 0cm},clip]{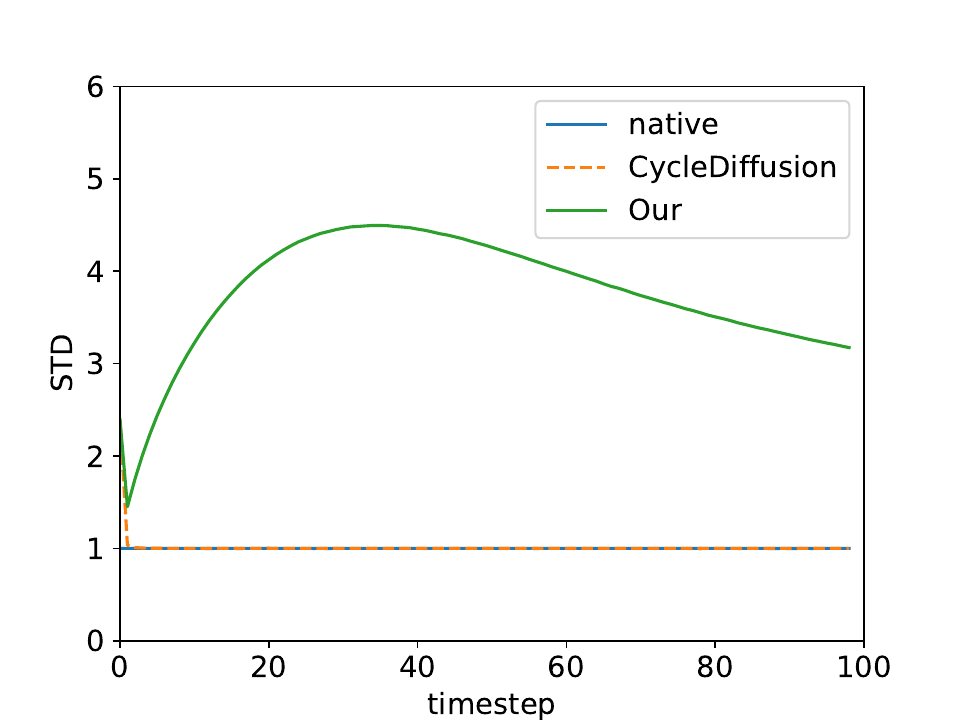}
\includegraphics[width=0.515\textwidth, trim={0cm 0cm 1.2cm 0cm},clip]{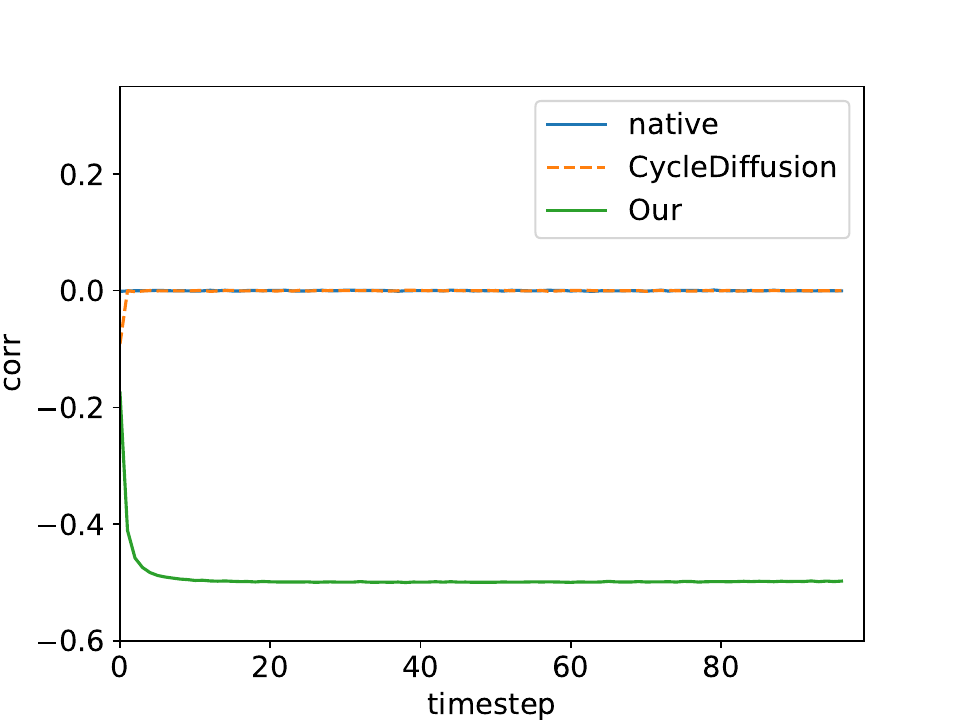}
\caption{\textbf{CycleDiffusion noise statistics.} Here we show the per-pixel standard deviations of $\{z_t\}$ and the per-pixel correlation between them for mssodel-generated images.}
\label{fig:statistics_images_CycleDiffusion}
\end{figure}

As mentioned in Sec.~\ref{sec:properties}, CycleDiffusion~\cite{Wu22} extracts a sequence of noise maps $\{x_T,z_T,\ldots,z_1\}$ for the DDPM scheme. However, in contrast to our method, their noise maps have statistical properties that resemble those of regular sampling. 
This is illustrated in Fig.~\ref{fig:statistics_images_CycleDiffusion}, which depicts the per-pixel standard deviations of $\{z_t\}$ and the correlation between $z_t$ and $z_{t-1}$ for CycleDiffusion, for regular sampling, and for our approach. These statistics were calculated over 10 images using an unconditional diffusion model trained on Imagenet, with $\eta=1.0$, $\text{strength}=3$ and $T_{\text{skip}} = 30$ as hyper parameters.
As can be seen, the CycleDiffusion curves are almost identical to those of regular sampling, and are different from ours. 

\begin{figure}[H]
\centering
\includegraphics[width=0.75\textwidth]{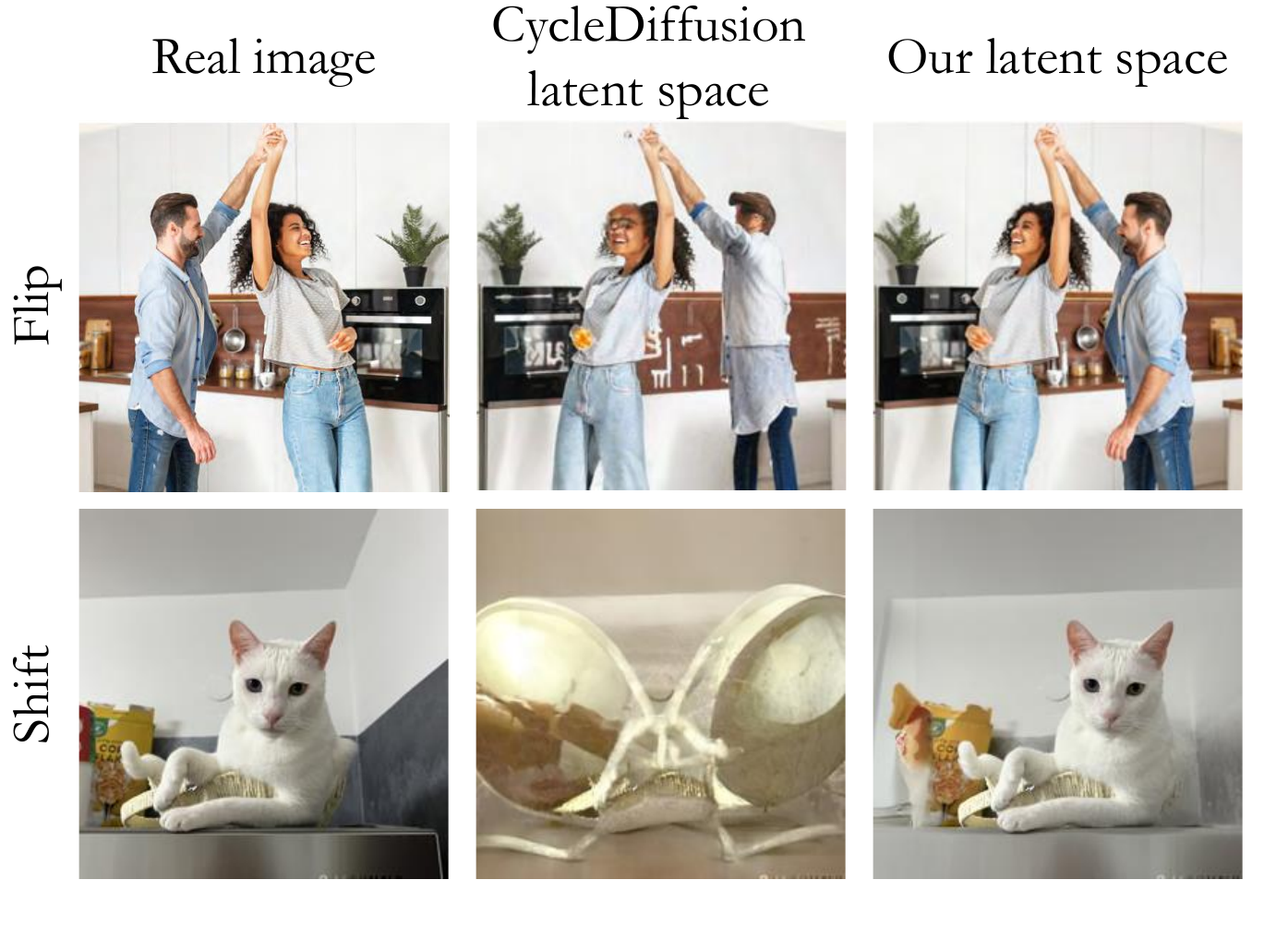}
\caption{\textbf{Flip and shift with CycleDiffusion and with our inversion.}}
\label{fig:flip_shift_CycleDiffusion}
\end{figure}

\vspace{-4.cm}
The implication of this is that similarly to the native latent space, simple manipulations on CycleDiffusion's noise maps cannot be used to obtain artifact-free effects in pixel space. This is illustrated in Fig.~\ref{fig:flip_shift_CycleDiffusion} in the context of horizontal flip and horizontal shift by 30 pixels to the right. As opposed to Cycle diffusion, applying those transformations on our latent code, leads to the desired effects, while better preserving structure.

\vspace{-0.2cm}
This behavior also affects the text based editing capabilities of CycleDiffusion. In particular, the CLIP similarity and LPIPS distance achieved by CycleDiffusion on the modified ImageNet-R-TI2I dataset are plotted in Fig.~\ref{fig:clip_lpips}. As can be seen, when tuned to achieve a high CLIP-similarity (\ie to better conform with the text), CycleDiffusion's LPIPS loss increases significantly, indicating that the output images become less similar to the input images. For the same level of CLIP similarity, our approach achieves a substantially lower LPIPS distance.

\clearpage

\section{The effect of skip and strength parameters}\label{app:SkipAndStrength}
\begin{figure*}[h]
\includegraphics[width=\textwidth]{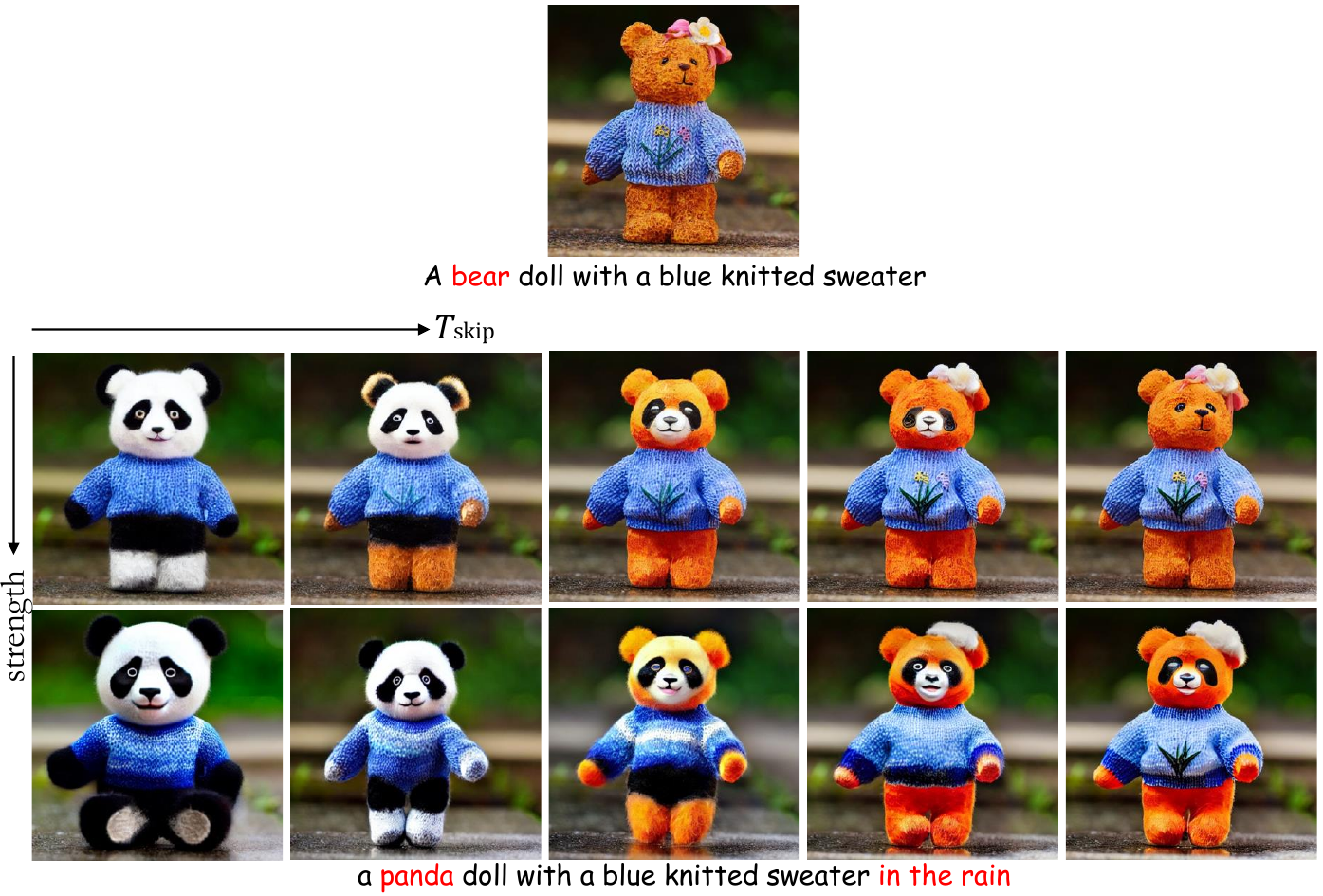}
\caption{\textbf{The effects of the skip and the strength parameters}.}
\label{fig:skip_strength}
\end{figure*}

Recall from Sec.~\ref{sec:text-guided} that in order to perform text-guided image editing using our inversion, we start by extracting the latent noise maps while injecting the source text into the model, and then generate an image by fixing the noise maps and injecting a target text prompt. Two important parameters in this process are  $T_{\text{skip}}$, which controls the timestep ($T-T_{\text{skip}}$) from which we start the generation process, and the strength parameter of the classifier-free scale~\cite{Ho21}.  Figure~\ref{fig:skip_strength} shows the effects of these parameters. When $T_{\text{skip}}$ is large, we start the process with a less noisy image and thus the output image remains close to the input image. On the other hand, the strength parameter controls the compliance of the output image with the target prompt.

\clearpage

\section{Integrating to P2P}
\begin{figure}
\centering
\includegraphics[width=0.65\textwidth]{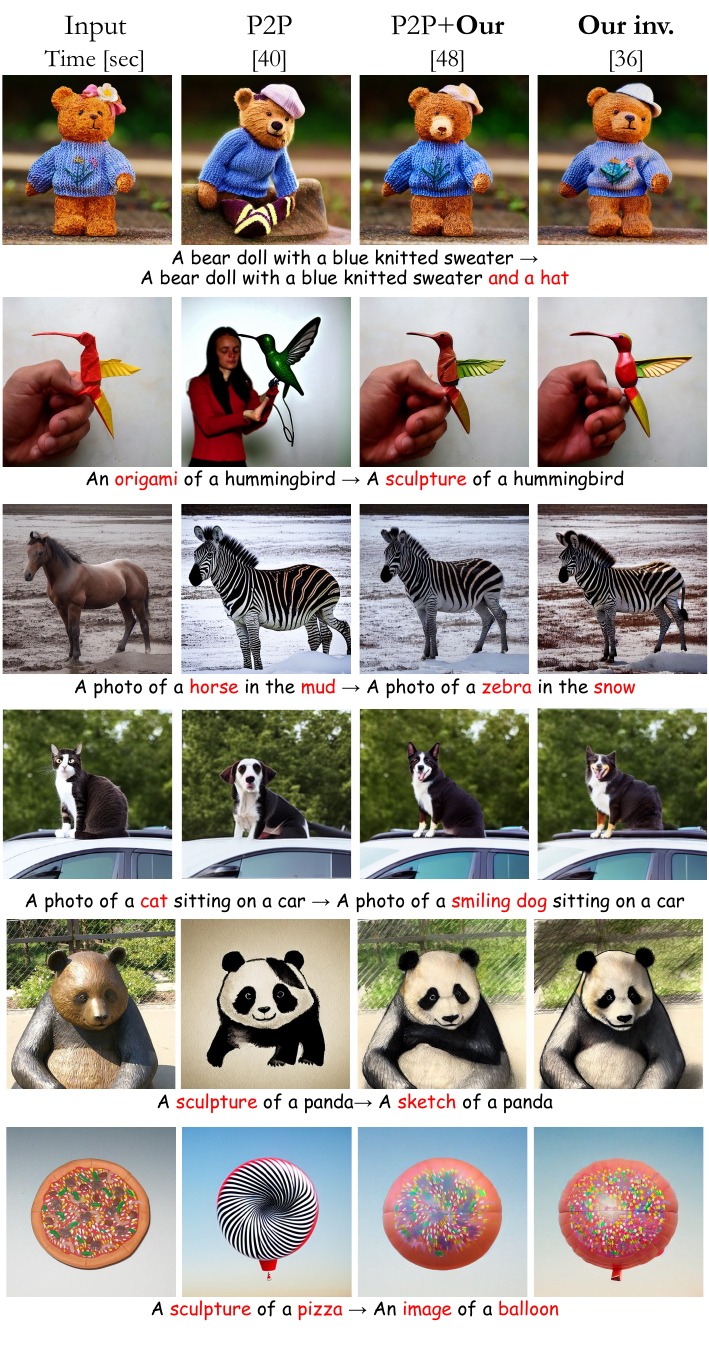}
\caption{\textbf{Comparison to P2P, with and without our inversion}. }
\label{fig:p2p_1}
\end{figure}

\begin{figure}
\centering
\includegraphics[width=0.65\textwidth]{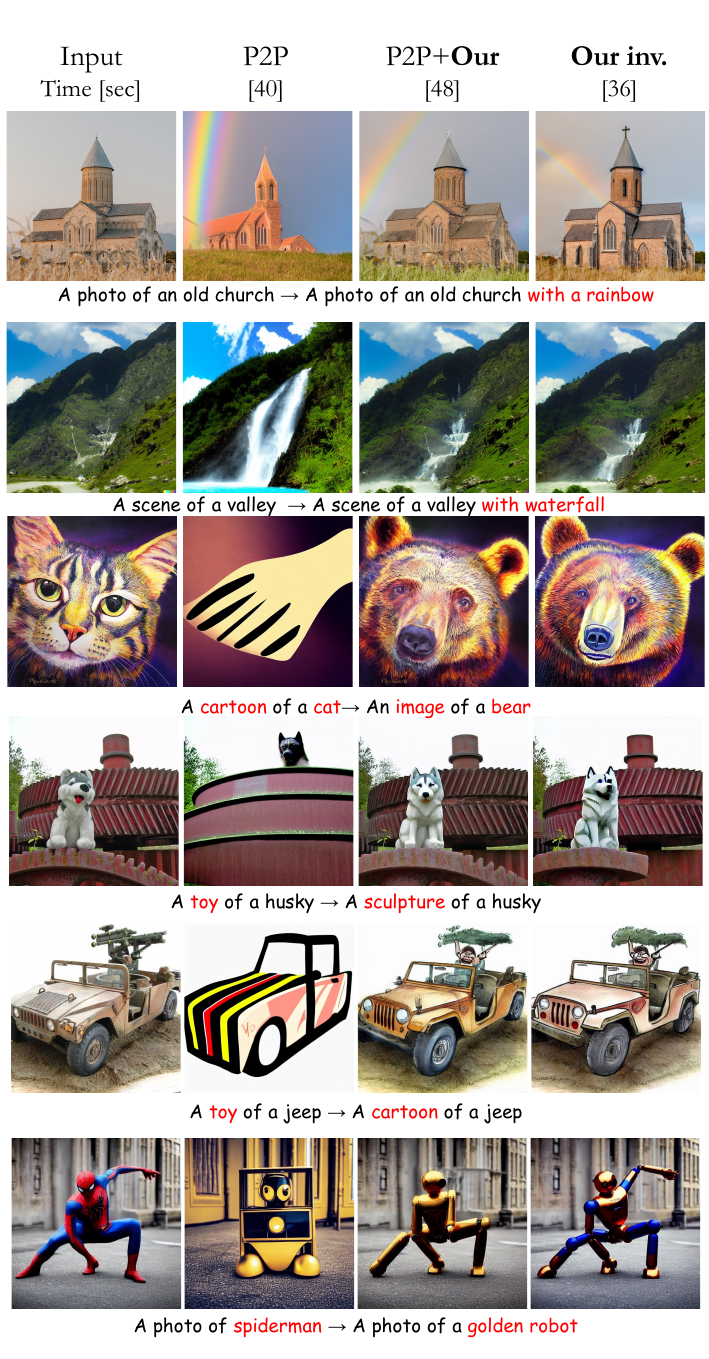}
\caption{\textbf{Additional comparisons to P2P, with and without our inversion}. }
\label{fig:p2p_2}
\end{figure}

As described in sec.~\ref{sec:text-guided}, our inversion method can be integrated with existing editing methods that rely on DDIM inversion. In addition to combining our method with Zero-Shot I2I, we assess the integration with Prompt-to-Prompt (P2P)~\cite{Hertz22}. In that case, we decrease the hyper-parameter controlling the cross-attention from $0.8$ to $0.6$ (as our latent space already strongly encodes structure). We note that P2P has different modes for different tasks (swap word, prompt refinement), and we chose its best mode for each image-prompt pair. Figure~\ref{fig:p2p_1} and ~\ref{fig:p2p_2} show that P2P does not preserve structure well. Yet, P2P does produce appealing results when used with our inversion. 

\clearpage

\clearpage

\clearpage
\section{Additional details on experiments and further numerical evaluation} 
\label{app:AdditionalDetails}
For all our text-based editing experiments, we used Stable Diffusion as our pre-trained text-to-image model. We specifically used the StableDiffusion-v-1-4 checkpoint. We ran all experiments on an RTX A6000 GPU. We now provide additional details about the evaluations reported in the main text. All datasets and prompts will be published.

In addition to measuring CLIP-based scores, LPIPS scores, and running time, we also measure the diversity among generated outputs (higher is better). Specifically, for each image and source text $p_{\text{src}}$, we generate 8 outputs with target text $p_{\text{tar}}$ and calculate the average LPIPS distance over all $\begin{psmallmatrix}8\\2\end{psmallmatrix}$ pairs.

\subsection{Experiments on the modified ImageNet-R-TI2I}\label{app:ImageNet-Dataset}
Our modified ImageNet-R-TI2I dataset contains 44 images: 30 taken from PnP~\cite{Narek22}, and 14 from the Internet and from the code bases of other existing text-based editing methods. We verified that there is a reasonable source and target prompt for each image we added. For P2P~\cite{Hertz22} (with and without our inversion), we used the first 30 images with the ``replace'' option, since they were created with rendering and class changes. That is, the text prompts were of the form ``a $\ll$rendering$\gg$ of a $\ll$class$\gg$'' (\eg ``a sketch of a cat'' to ``a sculpture of a cat''). The last 14 images include prompts with additional tokens and different prompt lengths (\eg changing ``A photo of an old church'' to ``A photo of an old church with a rainbow''). Therefore for those images we used the ``refine'' option in P2P. We configured all methods to use 100 forward and backward steps, except for PnP whose supplied code does not work when changing this parameter. 

Table~\ref{tab:table_parameter} summarizes the hyper-parameters we used for all methods. These apply to both the numerical evaluations and to the visual results shown in the figures. For our inversion, for P2P with our inversion, and for CycleDiffusion we arrived at those parameters by experimenting with various sets of parameters and choosing the configuration that led to the best CLIP loss under the constraint that the LPIPS distance does not exceed $0.3$. For DDIM inversion and for P2P (who did not illustrate their method on real images), such a requirement could not be satisfied. Therefore for those methods we chose the configuration that led to the best CLIP loss under the constraint that the LPIPS distance does not exceed $0.62$. We show results over DDIM inversion with 100 and 50 number of diffusion steps. For DDIM inversion mid-way we use the inversion until a specific timestep. For PnP, null-text inversion, and EDICT we used the default parameters supplied by the authors.


\begin{table}[H]
\centering
\small
\begin{tabular}{| c | c c | c c c|} 
 \hline
 \multirow{2}*{Method} & $\#$inv. & $\#$edit &  \multirow{2}*{strength} &  \multirow{2}*{$T_{\text{skip}}$}&  \multirow{2}*{$\tau_{\text{x}} \slash \tau_{\text{a}}$} \\
&steps & steps&& &

 \\[0.5ex]
 \hline\hline
  DDIM inv. $(T=100)$ &100 & 100 & 9& 0 & -- \\
  DDIM inv. $(T=50)$ &50 & 50 & 9& 0 & -- \\
P2P & 100& 100 & 9 & 0 & 80\slash40 \\ 
P2P + Our inv. &100 &100  & 9 & 12 & 60\slash20\\ 
 PnP & 1000 & 50 & 10 & 0 & 40\slash25 \\ 
 EDICT & 50 & 50 & 3 & 10 & --- \\ 
null-text inversion & 50 & 50 & 7.5 & 0 & 80\slash40 \\ 
 CycleDiffusion $(\eta=0.1)$ & 100 & 100 & 3 & 30 & -- \\  
 CycleDiffusion $(\eta=1.0)$ & 100 & 100 & 3 & 30 & -- \\  
Our inv. & 100 & 100 & 15 & 36 & --\\

\hline
\end{tabular}
\caption{\textbf{Hyper-parameters used in experiments on the modified ImageNet-R-TI2I dataset}. The parameter `strength' refers to the classifier-free scale of the generation process. As for the strength used in the inversion stage, we set it to $3.5$ for all methods except for PnP and CycleDiffusion which uses $1$. The timestep at which we start the generation is $T-T_{\text{skip}}$ and, in case of injecting attentions, we also report the timesteps determine until which step (starting from zero) the cross- and self-attentions are injected, $\tau_{\text{x}}$ and $\tau_{\text{a}}$ respectively.}
\label{tab:table_parameter}
\end{table}

\begin{figure}[h]
\centering
\includegraphics[width=0.7\columnwidth]{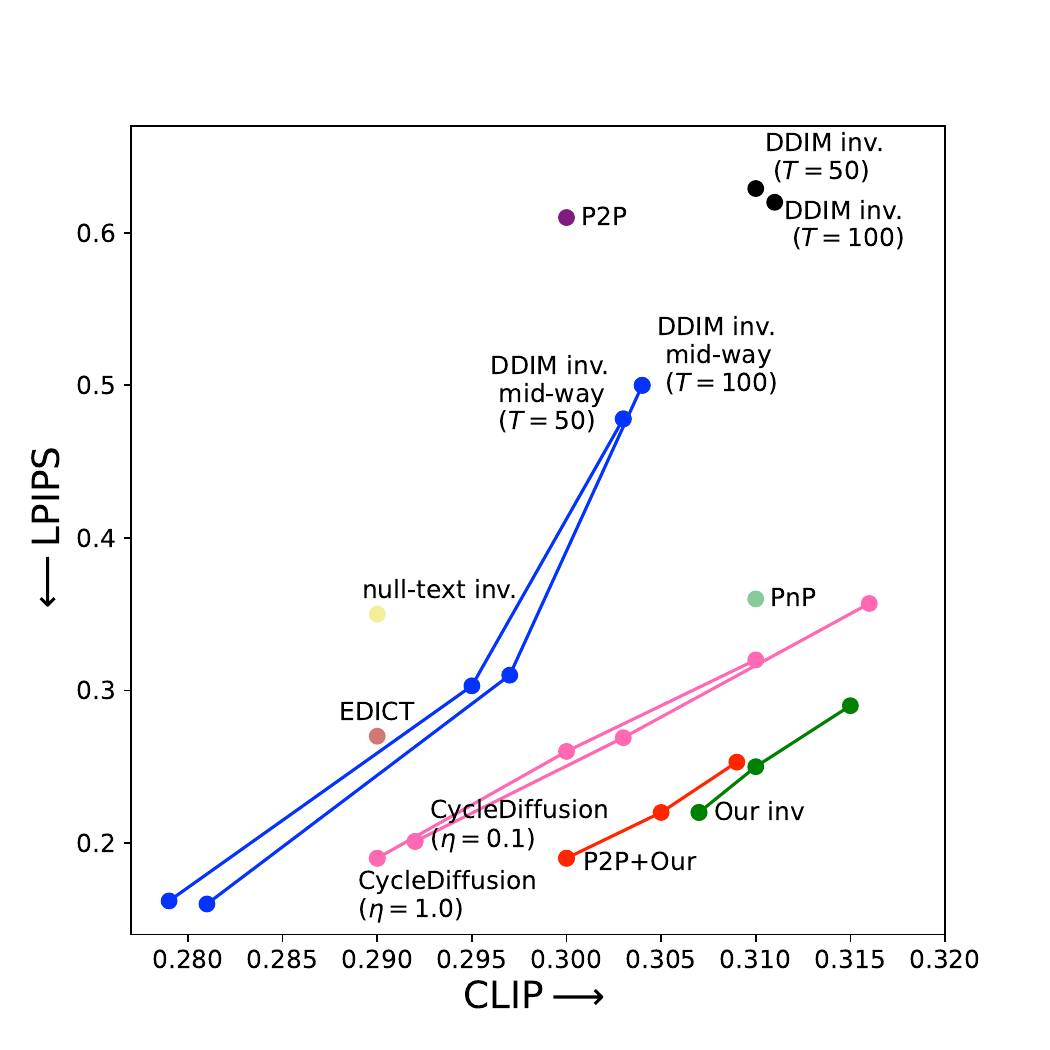}
\caption{\textbf{Fidelity to source image vs.~compliance with target text.} We show a comparison of all methods.}
\label{fig:clip_lpips_all}
\end{figure}

Table~\ref{tab:comparison_table} and Figure~\ref{fig:clip_lpips_all} summarizes the comparisons of all methods reported in the paper with the hyper-parameters from Tab~\ref{tab:table_parameter}. The results show that our inversion achieves a good balance between LPIPS and CLIP, while requiring short edit times. Integrating our inversion into P2P improves their performance in both metrics. Our method, CycleDiffusion, and null-text inversion support diversity among generated outputs.

\begin{table}[H]
\centering
\footnotesize
\begin{tabular}{| c | c c c c|} 
 \hline
 Method & CLIP sim.$\uparrow$ & LPIPS$\downarrow$ & Diversity$\uparrow$ & Time$\downarrow$ \\[0.5ex]
 \hline\hline
 DDIM inv. $(T=100)$  & 0.31 & 0.62 & 0.00 & 39\\
 DDIM inv. $(T=50)$  & 0.31 & 0.62 & 0.00 & 39\\
 P2P & 0.30 & 0.61 & 0.00 & 40\\
 \bf{P2P+Our}  & 0.31 & \bf{0.25} & 0.11 & 48\\  
 PnP & 0.31 & 0.36 & 0.00 & 206\\ 
 EDICT & 0.29 & 0.27 & 0.00 & 520\\ 
 null-text inversion & 0.29 & 0.35 & 0.08 & 160\\ 
 CycleDiffusion, \!$\eta\!=\!0.1$ & 0.30 & 0.27 & 0.21 &  \bf{36}\\ 
 CycleDiffusion, \!$\eta\!=\!1.0$ & 0.30 & 0.26 & \bf{0.306} &  \bf{36}\\ 
\bf{Our inv.}  & \bf{0.32} &0.29 & 0.18 & \bf{36}\\
\hline
\end{tabular}
\caption{Evaluation on modified ImageNet-R-TI2I dataset.}
\label{tab:comparison_table}
\end{table}


\clearpage

\subsection{Experiments on the modified zero-shot I2IT dataset}\label{app:zero-shot-dataset}
The second dataset we used is the modified Zero-Shot I2IT dataset, which contains 4 categories (cat, dog, horse, zebra). Ten images from each category were taken from Parmar \etal~\cite{Parmar23}, and we added 5 more images from the Internet to each category. Zero-Shot I2I~\cite{Parmar23} does not use source-target pair prompts, but rather pre-defined source-target classes (\eg cat$\leftrightarrow$dog). For their optimized DDIM-inversion part, they use a source prompt automatically generated with BLIP~\cite{Junnan22}. When combining our inversion with their generative method, we use $T_{\text{skip}}=0$ and an empty source prompt. Table~\ref{tab:table_parameter_zero_shot} summarizes the hyper-parameters used in every method.


\begin{table}[H]
\centering
\small
\begin{tabular}{| c | c c | c c c|} 
 \hline
 \multirow{2}*{Method} & $\#$inv. & $\#$edit &  \multirow{2}*{strength} &  \multirow{2}*{$T_{\text{skip}}$}&  \multirow{2}*{$\lambda_{\text{xa}}$} \\
&steps & steps&& &

 \\[0.5ex]
 \hline\hline
 Zero-Shot & 50 & 50 & 7.5 & 0 & 0.1 \\ 
Zero-Shot+Our &50 & 50 & 7.5 & 0 & 0.03 \\
\hline
\end{tabular}
\caption{\textbf{Hyper-parameters used in experiments on the modified Zero-Shot I2IT dataset}. In this method, cross-attention guidance weight is the parameter used to control the consistency in the cross-attention maps, denoted here as $\lambda_{\text{xa}}$. We set the strength (classifier-free scale) in the inversion part to be $1$ and $3.5$ for ``Zero-shot'' and ``Zero-shot+Our'' respectively.}
\label{tab:table_parameter_zero_shot}
\end{table}

Tab~\ref{tab:comparison_table_zero}
summarizes the comparison to the Zero-shot method. The results show that integrating our inversion improves the 
similarity to the input image while keeping the CLIP accuracy high. We also exhibit non-negligible diversity among the generated outputs

\begin{table}[H]
\centering
\footnotesize
\begin{tabular}{|c | c c c c|} 
 \hline
 Method & CLIP Acc.$\uparrow$ & LPIPS$\downarrow$ & Diversity$\uparrow$ & Time\\
\hline\hline
Zero-Shot  & 0.88 & 0.35 & 0.07 &\bf{45}\\ 
\bf{Zero-Shot+Our} & \bf{0.88} & \bf{0.27} & \bf{0.16} & 46\\  
\hline
\end{tabular}
\caption{Evaluation on the modified Zero-Shot I2IT dataset.}
\label{tab:comparison_table_zero}
\end{table}
\clearpage
\section{Additional results}

\begin{figure*}[h]
\includegraphics[width=\textwidth]{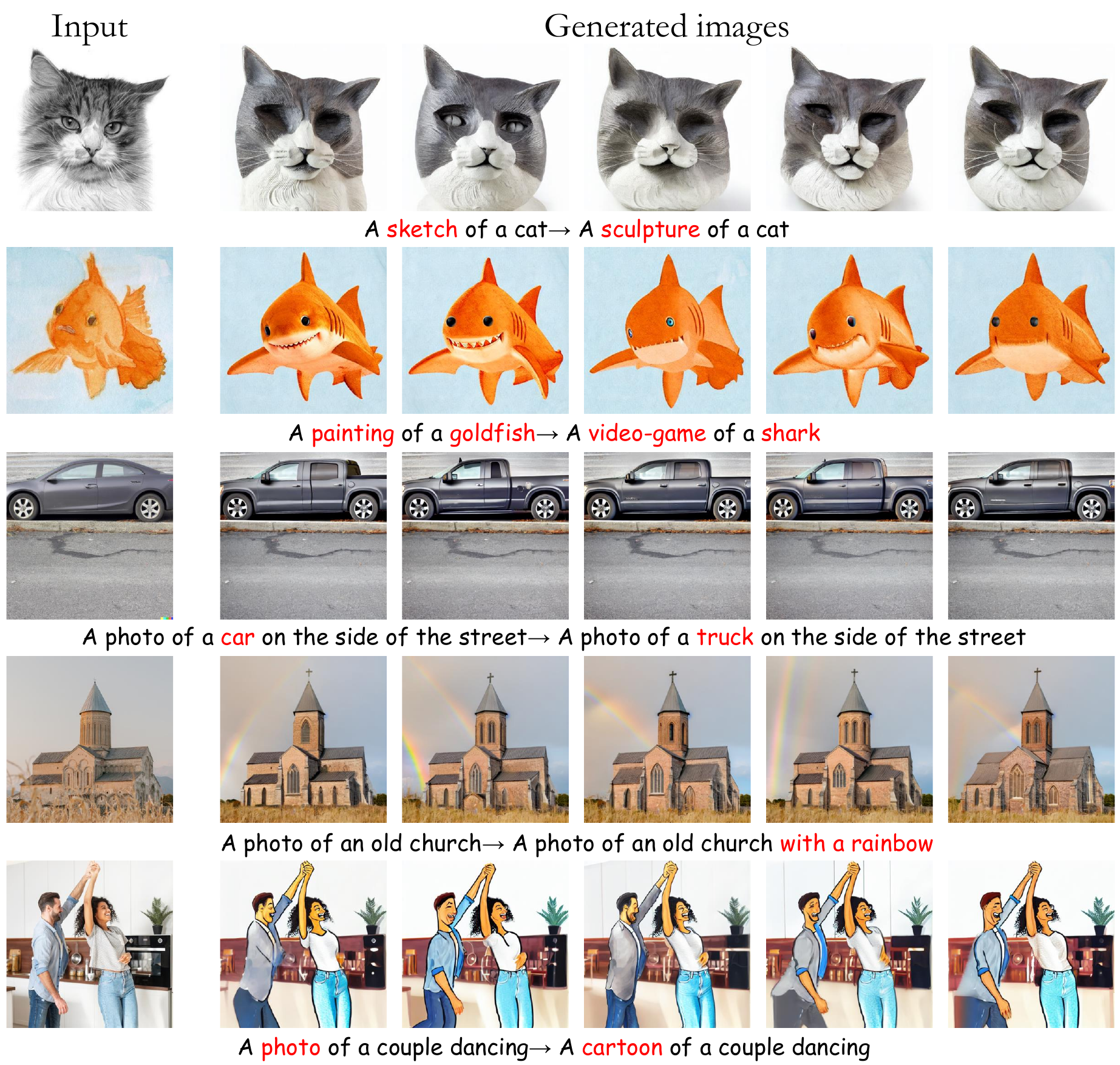}
\caption{\textbf{Diverse text-based editing with our method}. We apply our inversion five times with the same source and target prompts (shown beneath each example). Note how the variability between the results is not negligible, while all of them conform to the structure of the input image and comply with the target text prompt. Notice \eg the variability in the sculpture cat's eyes and mouth, and how the rainbow appears in different locations and angles.}
\label{fig:variability1}
\end{figure*}

\begin{figure*}
\includegraphics[width=\textwidth]{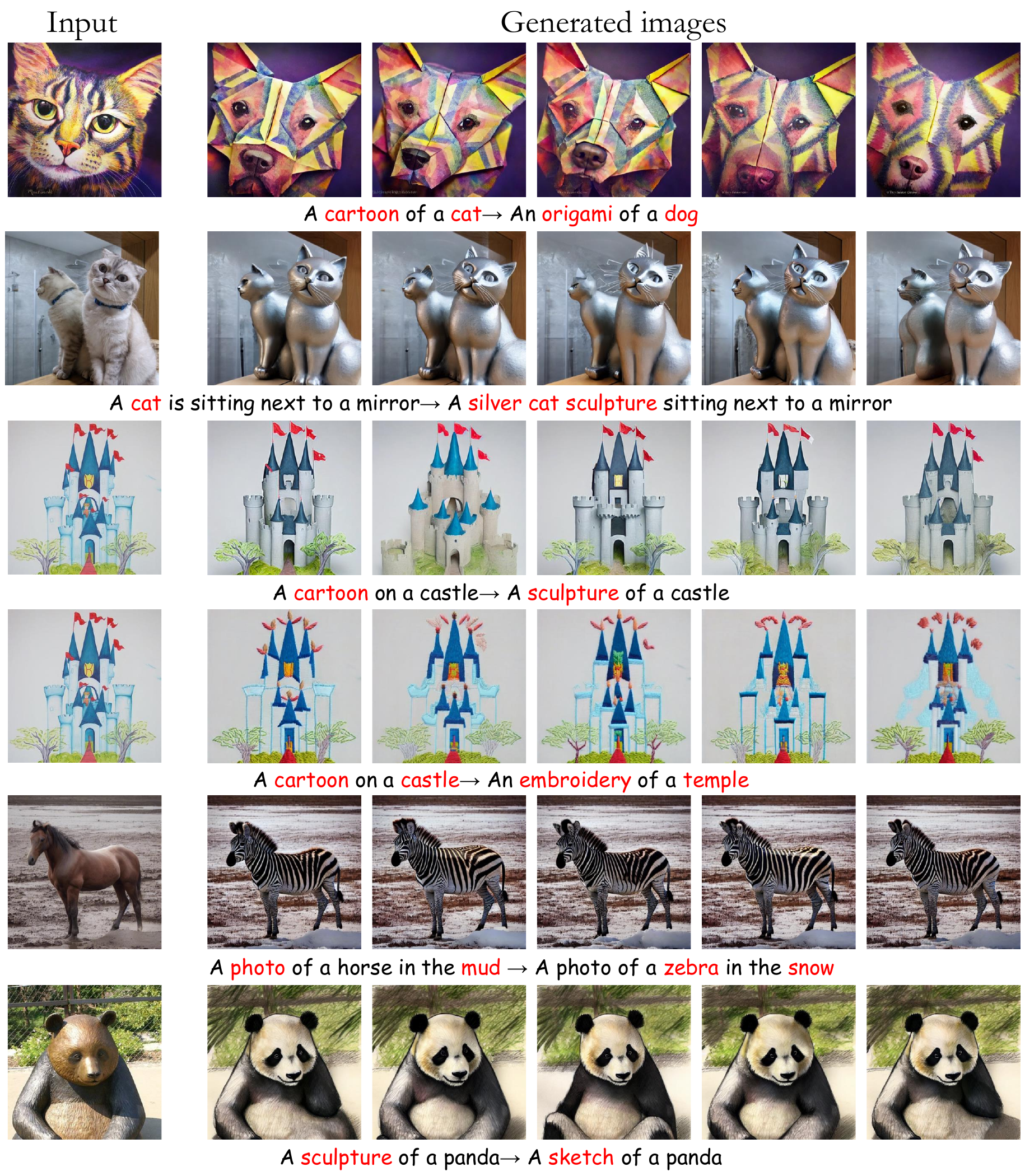}
\caption{\textbf{Additional results for diverse text-based editing with our method}. Notice that 
each edited result is slightly different. For example, the eyes and nose of the origami dog change between samples, and so do the zebra's stripes.}
\label{fig:variability2}
\end{figure*}

\begin{figure*}[h]
\includegraphics[width=\textwidth]{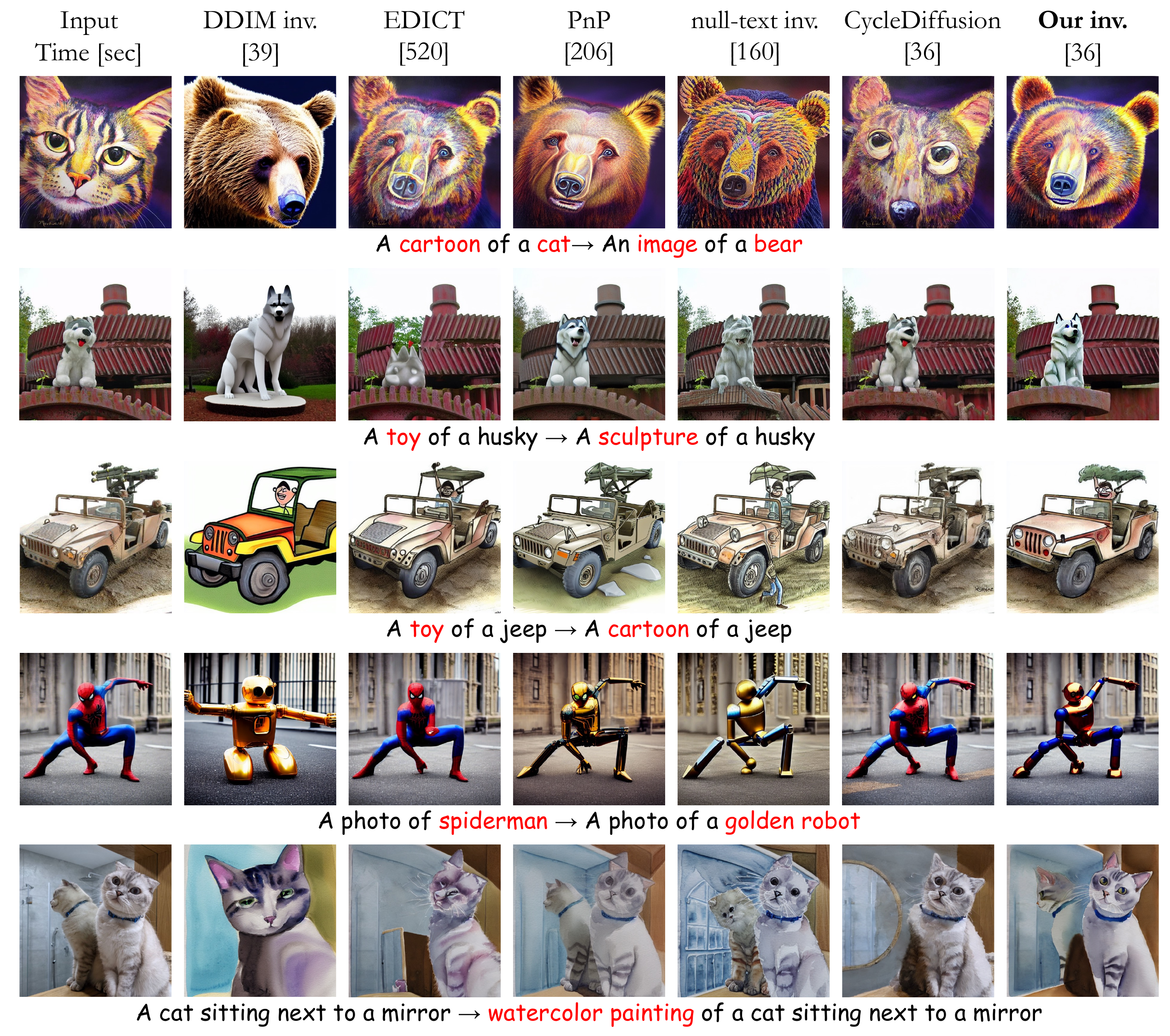}
\caption{\textbf{Qualitative comparisons between all methods.}}
\label{fig:comparisons_SM_1}
\end{figure*}

\begin{figure*}[h]
\includegraphics[width=\textwidth]{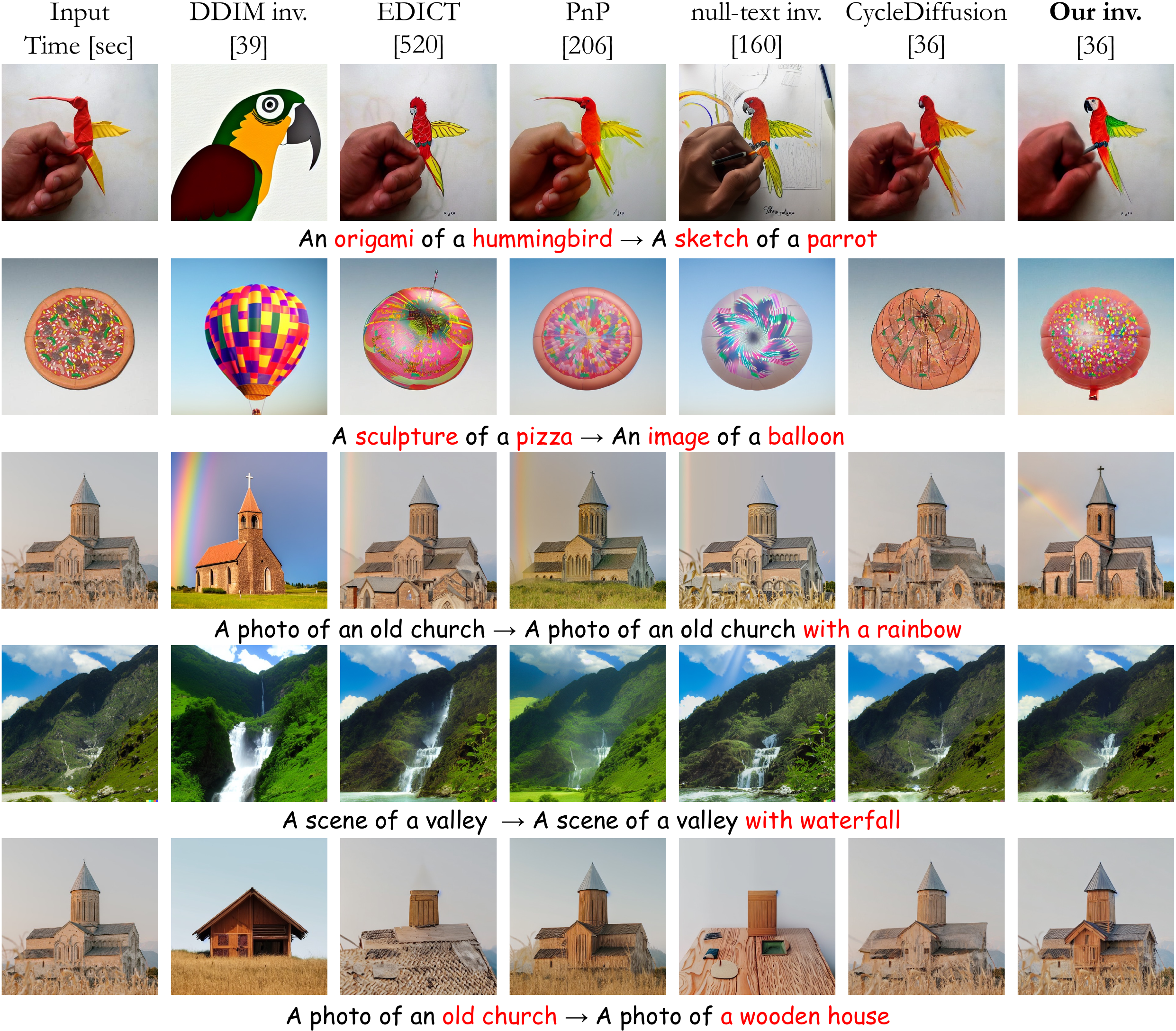}
\caption{\textbf{Additional qualitative comparisons between all methods.}}
\label{fig:comparisons_SM_2}
\end{figure*}

Due to the stochastic nature of our method, we can generate diverse outputs, a feature that is not naturally available with methods relying on the DDIM inversion. Figures~\ref{fig:variability1} and~\ref{fig:variability2} show several diverse text-based editing results.  Figures~\ref{fig:comparisons_SM_1} and \ref{fig:comparisons_SM_2} provide further qualitative comparisons between all methods tested on the ImageNet-R-TI2I dataset.

\clearpage

\clearpage



\end{document}